\documentclass[11pt,draftcls,onecolumn,journal]{IEEEtran}
\usepackage{etoolbox}
\AtBeginEnvironment{align}{\small}
\AtBeginEnvironment{align*}{\small}
\usepackage{amsmath,amsfonts}
\usepackage{amsthm}
\allowdisplaybreaks
\usepackage{algorithmic}
\usepackage{algorithm}
\usepackage{array}
\usepackage[caption=false,font=normalsize,labelfont=sf,textfont=sf]{subfig}
\usepackage{textcomp}
\usepackage{stfloats}
\usepackage{url}
\usepackage{verbatim}
\usepackage{graphicx}
\usepackage{cite}
\usepackage{hyperref}
\usepackage{booktabs}
\usepackage{enumitem}
\usepackage{xcolor}
\usepackage{tabularx}
\usepackage[most]{tcolorbox}

\usepackage[normalem]{ulem}

\newtheoremstyle{softdefinition}
  {6pt}
  {6pt}
  {\normalfont}
  {}
  {\itshape}
  {.}
  {.5em}
  {\thmname{#1}\thmnumber{ #2}\thmnote{ \normalfont(#3)}}

\theoremstyle{softdefinition}

\newtheorem{example}{Example}

\newtcolorbox{goalbox}{
  breakable,
  enhanced,
  colback=white,
  colframe=black,
  boxrule=0.5pt,
  arc=0pt,
  left=8pt,
  right=8pt,
  top=6pt,
  bottom=6pt,
  title={Learning goal},
  fonttitle=\bfseries
}

\begin{document}

\title{Causal Inference with Unobserved Confounding: A Mixture Learning Perspective}

\author{Mansi Sood and Devavrat Shah%
\thanks{Mansi Sood and Devavrat Shah are with the Laboratory for Information and Decision Systems, Massachusetts Institute of Technology, Cambridge, MA 02139 USA. E-mail: msood@mit.edu, devavrat@mit.edu. M.S. acknowledges support by the Schmidt Science Fellowship, in partnership with the Rhodes Trust. Both authors were supported in part by ONR MURI Project (N000142412687) titled Evaluating, Predicting, Optimizing, and Monitoring Interventions in Large Networked Systems.}}

\markboth{IEEE Signal Processing Magazine, Special Issue Article }%
{Sood and Shah: Causal Inference with Unobserved Confounding: A Mixture Learning Perspective}
\maketitle

 \vspace{-0.5in}
\begin{center}
\begin{minipage}{0.94\textwidth}
\small
\noindent\textbf{Abstract.}
{\bf
Unobserved confounding is a fundamental challenge in causal inference from observational data. This article develops a mixture-learning perspective, viewing latent confounders as sources of heterogeneity that induce mixture structure in observed data. Under suitable structural and identifiability assumptions, recovering the mixing distribution and component mechanisms enables estimation of interventional distributions and causal estimands. Using variants of Bernoulli mixtures as a running example, we contextualize mixture-learning techniques and their structural assumptions, and connect them to causal inference in panel-data settings, including latent factor models and synthetic interventions. We then consider high-dimensional exponential-family mixtures with dependent outcome trajectories, moving beyond counterfactual means to model counterfactual distributions. We situate this perspective relative to complementary approaches for unobserved confounding. Together, these ideas provide a bridge between mixture learning and causal inference, connecting recent advances in high-dimensional mixture learning to scalable identification and estimation of causal effects while raising new challenges for mixture learning.
}
\medskip
\begin{center}
\end{center}
\end{minipage}
\end{center}
\vspace{8pt}
\section*{Introduction}

\begin{figure}[t]
\centering
\includegraphics[width=0.92\linewidth]{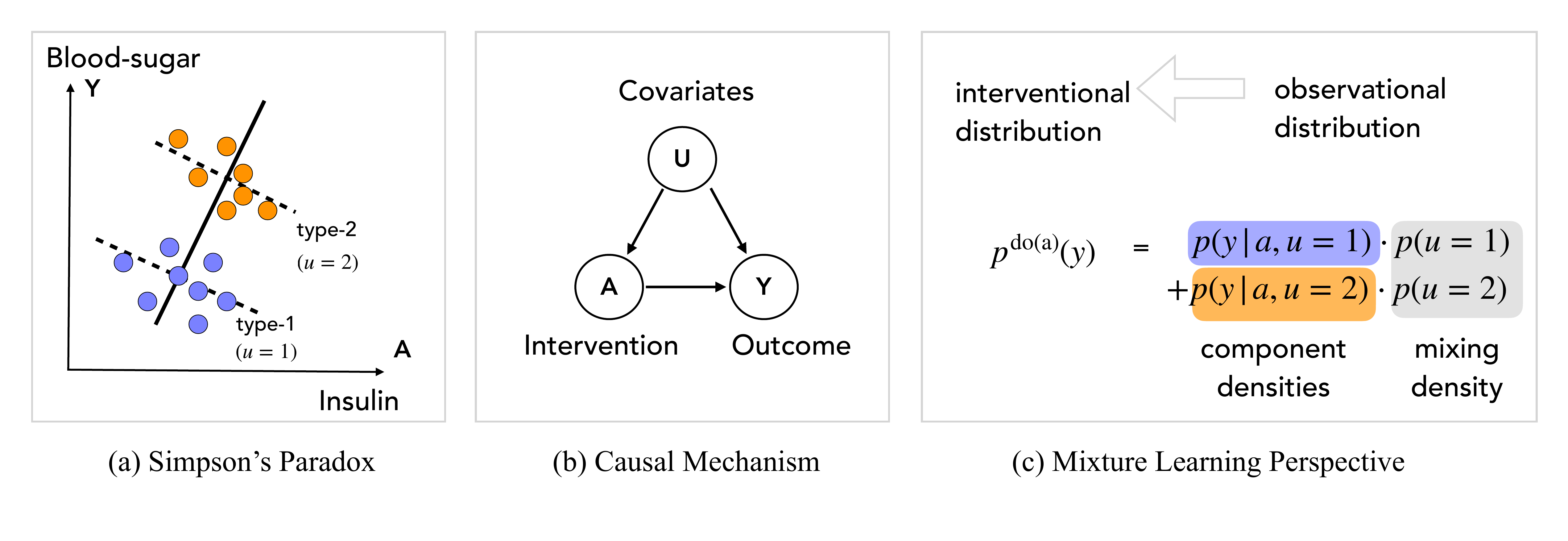}
\caption{An illustration of a mixture-learning-based resolution to Simpson's paradox. (a) The aggregate observational data (solid line) exhibits a spurious positive correlation between insulin dosage ($A$) and blood-sugar levels ($Y$). When conditioned on this latent confounder (shown as orange and blue clusters), the true negative causal effect is revealed (dashed lines). \textcolor{black}{(b) The causal mechanism is represented by a DAG
with latent confounder $U$.} (c) A mixture-learning perspective models unobserved confounding through a latent variable $U$ that induces a mixture distribution. By learning the confounding density $p(u)$ viewed as mixing density together with the component distributions $p(y\mid a,u)$, we can recover causal quantities from observational data under suitable identifiability conditions. }%
\label{fig:simpson-mixture}
\end{figure}

Causal inference is central to decision-making in high-stakes societal systems, where the goal is to quantify how outcomes \emph{change} under interventions rather than merely summarize correlations in observational data \cite{imbens2015causal}. Fundamentally, it seeks to answer the ``what if'' question: what would happen to an outcome of interest if a certain intervention were introduced? \textcolor{black}{An ideal approach to answering such questions} is
to conduct controlled randomized experiments and measure \textcolor{black}{the impact of an intervention} on the outcome of interest. Indeed, \textcolor{black}{the randomized controlled trial (RCT) protocols utilized by the Food and Drug Administration (FDA) to establish the safety and efficacy of medical products exemplify such an approach.} 
In many such settings, however, controlled experiments are infeasible, unethical, or prohibitively expensive, so one needs to instead rely on observational data. This is seen in the recent push for \textcolor{black}{the FDA to utilize real-world evidence from patients' electronic health records or other observational data \cite{Sherman2016, FDA_RWE_Portal,FDA_EHR_Guidance2024}.}  Indeed, 
the trace-driven simulators for engineering system design are following similar pursuits in complex, sequential settings \cite{Agarwal2021PerSim, Alomar2023CausalSim}. A key challenge in causal inference with observational data \textcolor{black}{is the potential for \emph{unobserved confounding}}, namely the presence of \emph{latent} factors that influence \emph{both} interventions and outcomes, thereby leading to incorrect causal estimates.
As an illustrative toy example, consider estimating the effect of insulin dosage on blood-sugar levels (Fig.~\ref{fig:simpson-mixture}). While physiology may suggest that higher insulin should reduce blood sugar, the aggregated observational data can misleadingly suggest \textcolor{black}{the opposite, a phenomenon known in the literature as Simpson's paradox} \cite{simpson1951}. This seeming paradox is explained by an unobserved confounder, namely diabetes type, which influences both the prescribed dosage and the insulin response. Specifically, by separating observational data into {\em mixture components} corresponding to \textcolor{black}{different diabetes types and then} \textcolor{black}{focusing on the relationship between insulin level and blood sugar level}, we observe the behavior that physiology would suggest. In summary, accounting for heterogeneity in the data \textcolor{black}{induced by the confounders} can lead to correct conclusions about causal effects.  

{
Concretely, accounting for such latent heterogeneity effectively means the following: \textcolor{black}{identify the key modes of heterogeneity and, for each observation, identify which mode of potentially latent heterogeneity is present.} \textcolor{black}{In the mixture learning literature}, the latent heterogeneity corresponds to the mixing component\textcolor{black}{,} and the primary goal is learning the mixing distribution \textcolor{black}{as well as the distribution of observations within each mixture component} \cite{EM-royal,linday-IMS}. {In summary}, given a causal mechanism \textcolor{black}{with a potentially discrete latent confounder (see} Fig.~\ref{fig:simpson-mixture}(b)),  the interventional distribution admits an explicit mixture representation, so learning the mixture components and mixing distribution enables causal estimation. While such a connection has been implicitly or explicitly utilized \textcolor{black}{in a case-by-case manner in the literature (see} \cite{gordon23a,saha,proximal,blei-deconfounder}\textcolor{black}{)}, the purpose of this overview article is to bring them together under a common umbrella as an important {\em bridge} so as to open up pathways for connecting recent advances in high-dimensional mixture learning to scalable identification and estimation of causal effects, while also raising new challenges for mixture learning.}


\noindent{\bf Organization.}
The remainder of the article is structured as follows. We begin with a primer on causal inference through the lens of graphical models. Specifically, we introduce notions of observational and interventional distributions and the {task of causal identification}. We then present a formal setup for causal inference under unobserved confounding and show how the resulting interventional distribution naturally takes a mixture form. Next, we discuss how learning the mixture distribution helps recover causal estimands \textcolor{black}{such as the average treatment effect}. \textcolor{black}{Using variants of Bernoulli mixtures as a running example, we contextualize mixture-learning techniques and their structural assumptions, and connect them to causal inference in panel-data settings, including latent factor models and synthetic interventions.} We then review recent advances in learning high-dimensional mixtures with dependent observations, particularly \textcolor{black}{in the context of exponential families}. Finally, we review complementary approaches to causal inference with unobserved confounding and discuss their relationship to the mixture-learning perspective developed in this article. We conclude by highlighting open directions for developing more flexible and scalable mixture-learning methods for causal estimation, including empirical validation of structural assumptions, continuous latent spaces, and overlapping mixture components.

\section*{\textcolor{black}{Graphical Models} and Causal Inference}


\noindent{\bf \textcolor{black}{Graphical Models.}} 
\textcolor{black}{At its core,} a probabilistic graphical model provides a succinct way to represent the 
joint distribution of a collection of random variables. Specifically, combinatorial structure in the graph captures conditional 
dependencies among the random variables. Typically, these models are \emph{undirected} (Markov random fields) or \emph{directed} (Bayesian networks)\footnote{\textcolor{black}{There is literature on graphical models involving} both directed and undirected edges. We shall refrain from such digressions.}. In the context of causal inference, our primary focus is on directed graphical models { represented via directed acyclic graphs (DAGs).}

An instance of a directed graph defines a family of distributions that factor according to functions of nodes and their parents. Consider
a directed graph $G=(\mathcal V,E)$ with ${\mathcal{V}} = \{1,\dots, N\}$\footnote{$N \geq 1$ is a positive integer.} 
corresponding to vertices and ${\mathcal{E}} \subset {\mathcal{V}} \times {\mathcal{V}}$ corresponding to directed edges so that
there are no directed cycles in $G$. Let $X_{\mathcal V}=\{X_i\}_{i\in\mathcal V}$ be a collection of random variables indexed by the nodes ${\mathcal{V}}$ with each $X_i \in \mathcal{X}$ for some alphabet $\mathcal{X}$. 
For each node $i\in\mathcal V$, let $\pi_i = \{ j \in {\mathcal{V}}~{:}~ (j,i) \in {\mathcal{E}}\}$ 
denote the set of parents of node $i$ in $G$. The family of distributions consistent with $G$ factorizes as
\begin{align}
p_{X_{\mathcal{V}}}(x_{\mathcal V})
=
\prod_{i\in\mathcal V} p_{X_i | X_{\pi_i}}(x_i \mid x_{\pi_i}),
\end{align}
where $x_{\mathcal V} \in \mathcal{X}^N$, $p_{X_i | X_{\pi_i}}(\cdot |\cdot)$ represents the conditional distribution of $X_i$ given $X_{\pi_i}$ for
$i \in \mathcal{V}$. 

\begin{figure}[t]
\centering
\includegraphics[width=0.99\linewidth]{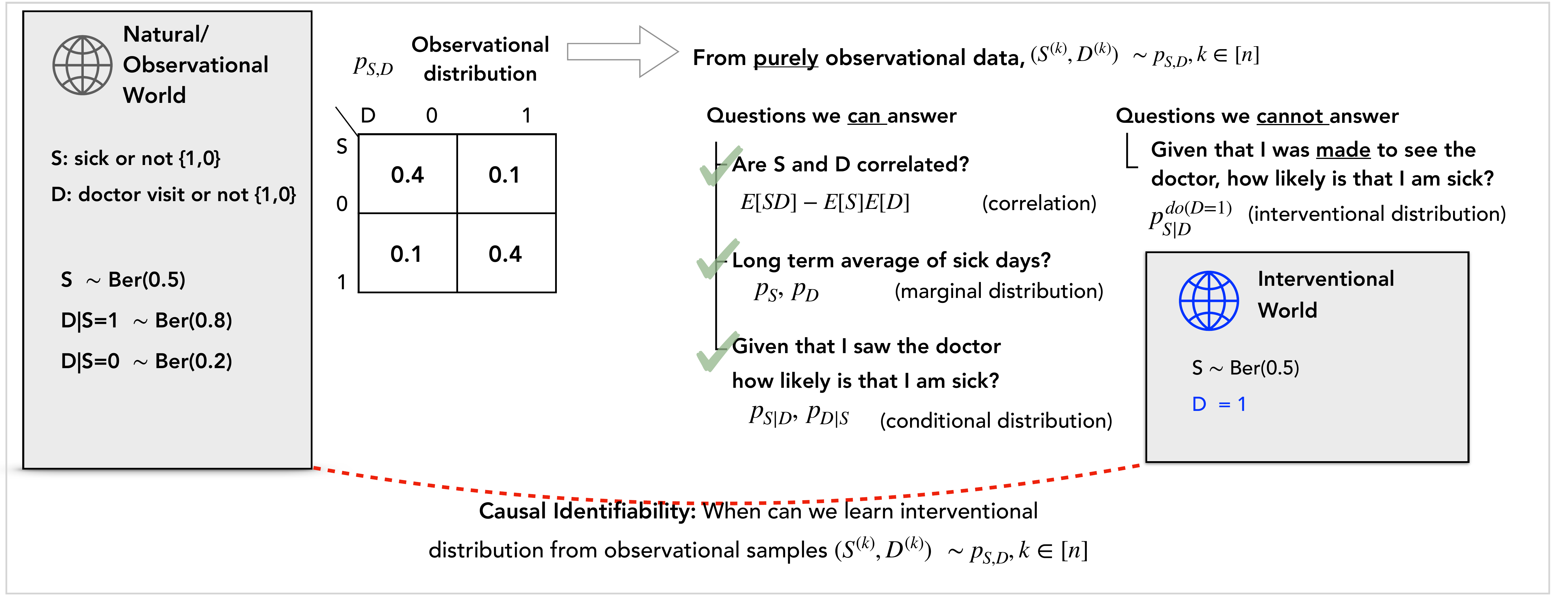}
\caption{Observational versus interventional worlds. While observational samples $(S^{(k)},D^{(k)})\sim p_{S,D}$ capture statistical associations between sickness ($S$) and doctor visits ($D$), they cannot determine interventional quantities, such as sickness outcomes under the intervention $\mathrm{do}(D=1)$. This gap motivates the causal identifiability question: determining the structural assumptions required to learn interventional distributions from purely observational data.}%
\label{fig:example-sick-doctor}
\end{figure}

\noindent{\bf \textcolor{black}{Observational vs. Interventional Distributions}: An Example.}
The \emph{observational distribution} is the distribution which describes the system in its natural state, 
that is, without any external intervention. The \emph{interventional distribution} is the distribution
describing the system when an intervention is applied to its natural state.
Let us introduce these notions first through an example. Suppose $S\in\{0,1\}$ represents whether I am sick today, and let $D\in\{0,1\}$ represent whether I go to the doctor tomorrow (Fig.~\ref{fig:example-sick-doctor}). 
Suppose we know that \textcolor{black}{the generative mechanism, or system in its natural state, behaves as follows:}
\begin{align}
S &\sim \mathrm{Ber}(0.5), &
D \mid S=1 &\sim \mathrm{Ber}(0.8), &
D \mid S=0 &\sim \mathrm{Ber}(0.2). \label{eq:sick}
\end{align}
That is, \textcolor{black}{every day I have a $50\%$ chance of being sick}, and if I am sick, then \textcolor{black}{I have an $80\%$ chance of visiting the doctor, and if I am not}
sick, then \textcolor{black}{I have a $20\%$ chance of visiting the doctor}. 
We {\em observe} whether I am sick or not, \textcolor{black}{i.e., $S$}, and whether I \textcolor{black}{go to the doctor or not, i.e., $D$}.  


The \emph{interventional} distribution describes the law induced after externally setting one or more variables to prescribed values while leaving the remaining causal mechanisms unchanged. In the example above, suppose I am made to go visit the doctor by being offered \$100 as an external incentive. This corresponds to the intervention of setting $D = 1$, which we shall represent as $\mathrm{do}(D=1)$ \textcolor{black}{and the resulting
joint distribution as} $p^{\mathrm{do}(D=1)}(\cdot)$, borrowing notation from
``do-calculus'' \cite{pearl2012calculus}. Under this intervention, the generating mechanism can be described as 
\begin{align}
S &\sim \mathrm{Ber}(0.5), \quad  
D \perp S, \quad D  \sim \mathrm{Ber}(1). \label{eq:sick.int}
\end{align}

The observational distribution of $(S, D)$, $p_{S, D}(\cdot)$, as per \eqref{eq:sick} can be \textcolor{black}{represented by a directed graphical model} $S\to D$ or $D \to S$. However, $S \to D$ captures {\em directionally} the underlying generative causal mechanism. The associated factor functions for $S \to D$
are $p_S(\cdot) \sim \mathrm{Ber}(0.5)$ and $P_{D|S}(\cdot | S=1) \sim \mathrm{Ber}(0.8), P_{D|S}(\cdot | S=0) \sim \mathrm{Ber}(0.2)$.  The interventional distribution of $(S, D)$ under intervention $\mathrm{do}(D=1)$, $p^{\mathrm{do}(D=1)}_{S, D}(\cdot)$, as per \eqref{eq:sick.int}, can be \textcolor{black}{represented by a directed
graphical model with no edges}. The corresponding factor functions are $p^{\mathrm{do}(D=1)}_S(\cdot) \sim \mathrm{Ber}(0.5)$ and $p^{\mathrm{do}(D=1)}_D(\cdot) \sim \mathrm{Ber}(1)$. 
\begin{figure}[t]
\centering
\includegraphics[width=0.83\linewidth]{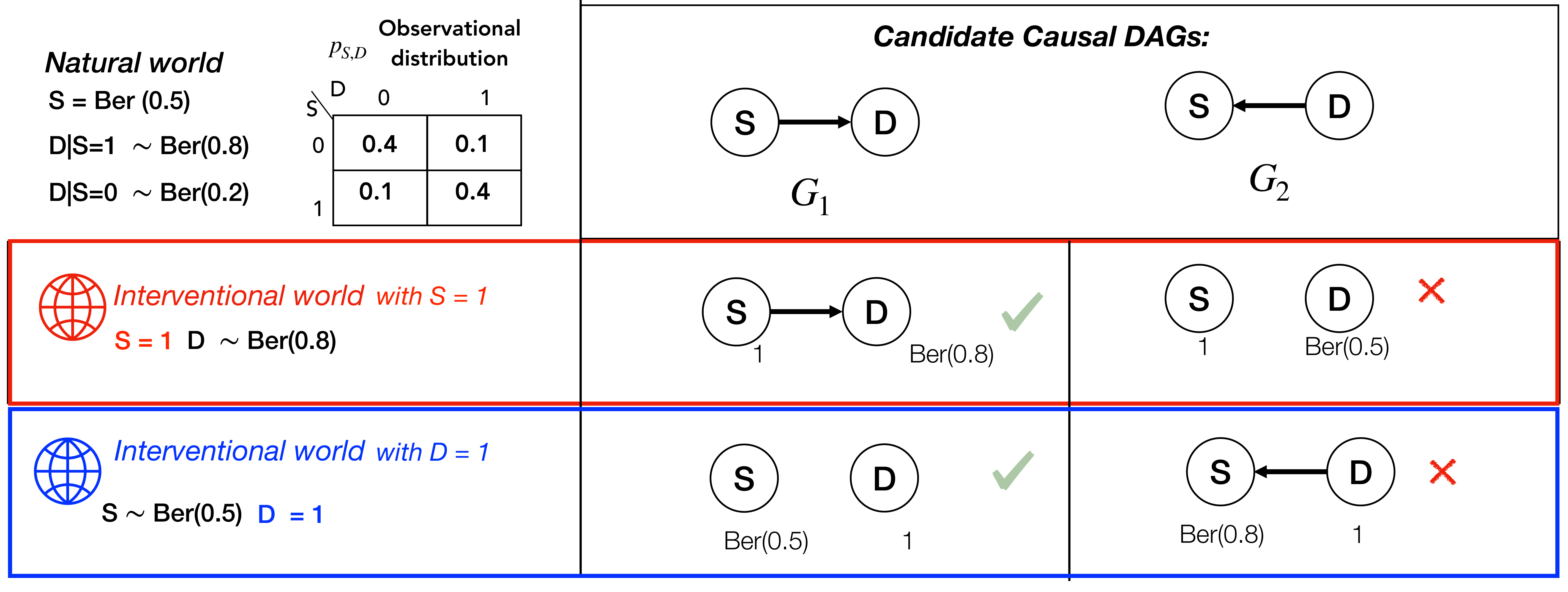}
\caption{Evaluating candidate causal DAGs via intervention axioms. \emph{(Top)} Observational data alone cannot distinguish between $G_1$ ($S \to D$) and $G_2$ ($S \gets D$). \emph{(Middle)} Under the intervention $\mathrm{do}(S=1)$, removing incoming edges to $S$ correctly preserves $D \sim \mathrm{Ber}(0.8)$ in $G_1$, but incorrectly isolates $D$ in $G_2$. \emph{(Bottom)} Under $\mathrm{do}(D=1)$, removing incoming edges to $D$ correctly isolates the marginal $S \sim \mathrm{Ber}(0.5)$ in $G_1$, while $G_2$ incorrectly predicts $S \sim \mathrm{Ber}(0.8)$.}
\label{fig:valid-causal-dags}
\end{figure}

\noindent{\bf \textcolor{black}{Observational vs. Interventional Distributions: Causal Mechanisms, Causal DAGs.}}
The example discussed illustrates that observational and interventional distributions are different. However, they do agree on the \textcolor{black}{aspects of the mechanism} 
that are unchanged under the intervention. And it may be possible to \textcolor{black}{``infer'' an interventional distribution from the observational distribution} when
we \textcolor{black}{have an understanding of the causal mechanism captured through the appropriate DAG, which we shall denote as the causal DAG}. \textcolor{black}{With this context}, 
\textcolor{black}{we discuss a generic causal mechanism, the corresponding causal DAG, and the relationship between the observational distribution and interventional distributions}. 

We start with \textcolor{black}{the description of a causal mechanism, also called a} \emph{structural causal model} (SCM), describing how the observational distribution is generated. 
As before, let $X_{\mathcal V}$ denote the collection of variables with $\mathcal{V} = \{1,\dots, N\}$. Let $\pi_i \subset \mathcal{V}$ \textcolor{black}{be the parents of node} $i$ for each $i \in \mathcal{V}$. 
Without loss of generality, we shall assume that for any $i, j \in \mathcal{V}$ if $j > i$ then $j \notin \pi_i$, \textcolor{black}{i.e., the labeling of variables} provides an order that is consistent with the
parent relationship. The $X_{\mathcal V}$ are generated as follows: for $i = 1,\dots, N$, generate $X_i$ sequentially as 
\begin{align}\label{eq:scm}
    X_i & =  f_i(X_{\pi_i},\varepsilon_i), 
\end{align}
where $\varepsilon_i, i \in \mathcal{V}$ are independent random variables. The equation \eqref{eq:scm} determines $p_{X_i | X_{\pi_i}}(\cdot | \cdot)$ for each $i \in \mathcal{V}$. It also \textcolor{black}{corresponds to
a directed graphical model} 
\begin{align}\label{eq:causalDAG}
    G & = (\mathcal{V}, \mathcal{E}), \quad \text{where}\quad \mathcal{E} = \cup_{i \in \mathcal{V}} \{ (j, i): j \in \pi_i\}.
\end{align}
The corresponding factor functions are precisely $p_{X_i | X_{\pi_i}}(\cdot | \cdot)$. In that sense, this DAG captures the causal mechanism exactly. We call it the causal DAG. Such a causal DAG naturally satisfies three key
properties, which will be stated soon, \textcolor{black}{that connect it} to any interventional distribution. Before we state them, we \textcolor{black}{formalize the notion of} interventional distributions. 

Let $I \subseteq \mathcal V$ and let $x_I' \in \mathcal{X}^{|I|}$ be an assignment of values to the variables $\{X_i\}_{i\in I}$. 
The interventional distribution under $\mathrm{do}(X_I=x_I')$ is the distribution induced by externally setting $X_I$ to $x_I'$ while 
leaving the remaining causal mechanisms unchanged. We denote it by
$
p_{X_{\mathcal V}}^{\mathrm{do}(X_I=x_I')}(x_{\mathcal V}).
$
Let
\begin{align}
 \mathcal{P}^*
=
\left\{
P_{X_{\mathcal V}}^{\mathrm{do}(X_I=x_I')}
:
I\subseteq\mathcal V,\,
x_I' \in \mathcal X^{|I|}
\right\}
\end{align}
denote the collection of all observational and interventional distributions, where the observational distribution corresponds to $I=\emptyset$.  
We say that $ \mathcal{P}^*$ is consistent with a \emph{causal DAG} $G$ if and only if, for every intervention $\mathrm{do}(X_I=x_I')$, the following hold:
\begin{enumerate}
    \item \emph{Factorization:} $P_{X_{\mathcal V}}^{\mathrm{do}(X_I=x_I')}$ factorizes with respect to $G$.
    \item \emph{Intervention:} for every $i\in I$, the marginal law of $X_i$ under $P_{X_{\mathcal V}}^{\mathrm{do}(X_I=x_I')}$ is a point mass at $x_i'$.
    \item \emph{Invariance:} for every $i\notin I$,
    \begin{align}
    P_{X_i\mid X_{\pi_i}}^{\mathrm{do}(X_I=x_I')}(x_i\mid x_{\pi_i})
    =
    P_{X_i\mid X_{\pi_i}}(x_i\mid x_{\pi_i}),
    \end{align}
    for all parent configurations $x_{\pi_i}$ consistent with the intervention values $x_I'$.
\end{enumerate}
It can be argued that \textcolor{black}{if the observational distribution is generated as per the causal mechanism} \eqref{eq:scm} and the DAG of interest $G$ is defined as per \eqref{eq:causalDAG}, \textcolor{black}{then it is indeed
{\em a causal DAG}} as it will satisfy the properties of factorization, intervention and invariance. 

Given these properties, any interventional distribution $P' \in \mathcal{P}^*$ can be obtained \textcolor{black}{by applying these properties with knowledge of the observational distribution represented by the causal DAG}. Such {\em calculus} is known as ``do-calculus'' \cite{pearl2012calculus}. The factorization condition dictates that interventions do not introduce new conditional dependencies beyond those already encoded by the graph. 
The intervention condition ensures that intervened variables are set precisely to their prescribed values. Finally, the invariance condition states that variables not directly intervened upon retain their original conditional mechanisms given their parents. 
Together, these conditions formalize the principle that an intervention replaces only the mechanisms at the targeted nodes, leaving the rest of the system intact. Graphically, this corresponds to a \emph{pruned} causal DAG, where all incoming edges to the intervened nodes are removed (Fig.~\ref{fig:valid-causal-dags}).

\noindent{\bf \textcolor{black}{Causal Estimands.}}
Typically, the quantity of interest is \textcolor{black}{a functional of an interventional distribution or, more precisely, a difference in the quantity of interest} under different interventional distributions. These are known as
{\em causal estimands}. As discussed above, if we have access to observational data and \textcolor{black}{knowledge of the causal DAG}, then in effect any reasonable causal estimand \textcolor{black}{boils down to the estimation of appropriate conditional distributions (corresponding to factors of the causal DAG)} and then utilizing ``do-calculus''. In most settings, the observational data may not 
necessarily reveal information about all relevant variables. And the causal mechanism driving the observational data or distribution may not be (fully) known or learnable.

This leads to the basic question in most settings: \textcolor{black}{{\em is the causal estimand identifiable?} That is, given the distribution of observed variables and the structural assumptions made, 
can the causal estimand be expressed uniquely as a functional of this distribution?} {\color{black}When identifiability fails, there exist two distinct causal models, both compatible with the structural assumptions, that induce the same observational distribution but imply different values for the target causal estimand.} In that case, observational data alone cannot determine the causal estimand without additional information and assumptions.


\section*{Causal inference under unobserved confounding}

We come to the 
core focus of this article: learning causal estimands from observational data with \emph{unobserved confounding}.~Towards that and without loss of generality, \textcolor{black}{we will simplify our notation}. Let $Y\in\mathcal Y$ denote the outcome variable of interest,  
$A\in\mathcal A$ denote the intervention or action or treatment variable, \textcolor{black}{and let all other relevant variables be denoted by} $U\in\mathcal U$. Each of these
variables can be high-dimensional. We assume that the joint distribution of $(U,A,Y)$ is generated by a causal DAG  ($U\to A$, $U\to Y$, and $A\to Y$) 
(Figure~\ref{fig:simpson-mixture} (b)) through the (simplified) structural model
\begin{align} 
U \;=\; \varepsilon_1,\qquad
A \;=\; f(U,\varepsilon_2),\qquad
Y \;=\; g(A,U,\varepsilon_3),
\label{eq:sem}
\end{align} 
where $\varepsilon_1,\varepsilon_2,\varepsilon_3$ are mutually independent exogenous variables; $f$ and $g$ are deterministic functions. 
We assume that, following the mechanism in \eqref{eq:sem}, $n$ independent samples $(u^{(k)}, a^{(k)}, y^{(k)})$ are generated for $k \in [n]$. 
Our overarching goal is to estimate causal quantities under interventions on the treatment variable. We shall assume that we do not 
observe $u^{(k)}$ but we do observe $(a^{(k)}, y^{(k)})$ for all $k \in [n]$.

\noindent Note that the distribution over $(U, A, Y)$ factorizes as
\begin{align}
p(u,a,y)
=
p(u)\,p(a\mid u)\,p(y\mid a,u),
\label{eq:obs-factorization}
\end{align}
and the observational marginal over $(A,Y)$ can be written as\footnote{For simplicity, throughout we shall assume that the alphabets $\mathcal{A}, \mathcal{U}, \mathcal{Y}$ are finite or
countably infinite.}
\begin{align}
p(a,y)
=
\sum_{u \in \mathcal{U}} p(u)\,p(a\mid u)\,p(y\mid a,u).
\label{eq:obs-mixture}
\end{align}

We call this setting {\em unobserved confounding}: $U$ {\em confounds} the intervention or action $A$ and outcome $Y$ in the observed data; and $U$ is {\em not observed}: we 
observe action-outcome pairs $(a^{(k)},y^{(k)})$ drawn per $p(a,y)$ \eqref{eq:obs-mixture} while the latent type $u^{(k)}$ remains hidden. 
This presents the key challenge: because the unobserved variable $U$ influences both treatment assignment and outcomes, differences in the observed outcome distribution across treatments do not solely reflect the causal effect of $A$ on $Y$. They also capture shifts in the latent composition of the population induced by $U$. Consequently, units receiving different treatments may systematically differ in their hidden characteristics, meaning the naive observational association between $A$ and $Y$ may not coincide with the true interventional effect of setting $A$ to a specific value. To make the role of confounding apparent, we consider two examples.

\begin{example}[\emph{Tobacco policy \cite{abadie2010synthetic,agarwal2025synthetic}}]\label{ex:tobacco-policy}
{As a motivating example, consider the adoption of anti-tobacco policies, including California's Proposition~99. In 1988, California passed Proposition~99, the first large-scale anti-tobacco program in the United States. Early reports of its effectiveness led to a wave of measures across the country between 1989 and 2000, with 4 states introducing similar anti-tobacco programs and 7 states increasing their cigarette taxes, while the remaining states retained the status quo. Let $A$ indicate whether a state has \textcolor{black}{an anti-tobacco program or tax in effect, or retains the status quo in a given year}, and let $Y$ denote tobacco consumption. The latent covariates $U$ represent state- and year-specific characteristics, such as socio-political conditions, demographics, and baseline trends, that affect both policy adoption and consumption. These hidden factors create confounding because they influence both treatment assignment and the outcome.}
\end{example}

\begin{example}[\emph{Information diffusion\cite{martin-johan}}]
{As another example, consider estimating the effect of attaching a Community Note to a social-media post on subsequent engagement and diffusion. A Community Note is a crowdsourced fact-checking feature primarily used \textcolor{black}{on the social media platform X} (formerly Twitter). Let $A\in\{0,1\}$ indicate whether a post receives a note, and let $Y$ denote an outcome that quantifies engagement, such as the number of reposts or views over a fixed time horizon. The latent covariates $U$ capture unobserved post-level characteristics, such as the post's intrinsic virality and audience susceptibility, that influence both the likelihood of receiving a note and the post's subsequent diffusion trajectory. These hidden factors create confounding because posts that are more likely to be flagged and annotated may also be precisely those that would have exhibited different growth trajectories even in the absence of the intervention.}
\end{example}

\noindent {\bf Key causal estimands.} Let $p^{\mathrm{do}(A=a')}(u,a,y)$ denote the joint distribution over $(U,A,Y)$ under the intervention $A=a'$, where other \textcolor{black}{aspects of the mechanisms} in \eqref{eq:sem} remain intact. Our primary causal quantity of interest is the expected outcome $\mathbb{E}^{\mathrm{do}(A=a')}[Y]$. En route, we aim to recover the marginal interventional distribution $p^{\mathrm{do}(A=a')}(y)$. Using the causal DAG (Figure~\ref{fig:simpson-mixture} (b)) and invoking the factorization, intervention, and invariance axioms, we can expand $p^{\mathrm{do}(A=a')}(y)$ as follows:
\begin{align}
p^{\mathrm{do}(A=a')}(y)
&= \sum_{\substack{u\in\mathcal U\\ a\in\mathcal A}} p^{\mathrm{do}(A=a')}(u,a,y) \nonumber \\
&= \sum_{\substack{u\in\mathcal U\\ a\in\mathcal A}} p^{\mathrm{do}(A=a')}(u)\, p^{\mathrm{do}(A=a')}(a\mid u)\, p^{\mathrm{do}(A=a')}(y\mid u,a) \label{eq:intervention-factorize} \\
&= \sum_{\substack{u\in\mathcal U\\ a\in\mathcal A}} p(u)\, p^{\mathrm{do}(A=a')}(a\mid u)\, p(y\mid u,a) \label{eq:intervention-invariance} \\
&= \sum_{u\in\mathcal U} p(u)\,p(y\mid u,a'). \label{eq:intervention-via-mixture}
\end{align}

This derivation proceeds in three steps. First, Equation \eqref{eq:intervention-factorize} applies the factorization axiom to the post-intervention distribution. Second, Equation \eqref{eq:intervention-invariance} uses the invariance axiom for the non-intervened nodes: because $U$ has no parents and is not intervened upon, its marginal law is unchanged; similarly, because $Y$ is not intervened upon, its conditional mechanism given $(U,A)$ remains the same. Finally, Equation \eqref{eq:intervention-via-mixture} relies on the intervention axiom, where the post-intervention distribution of $A$ collapses to a point mass at $a'$, leaving only the terms where $a=a'$.

It is important to distinguish the form of \eqref{eq:intervention-via-mixture} from the observed conditional distribution:
\begin{align}
p(y\mid a')
=
\sum_{u\in\mathcal U} p(u\mid a')\,p(y\mid a',u). \label{eq:compare-obs}
\end{align}
Comparing \eqref{eq:intervention-via-mixture} and \eqref{eq:compare-obs}, the difference is in the mixing weights: observational conditioning uses the confounder distribution conditioned on treatment $p(u\mid a')$, while intervention uses the population latent distribution $p(u)$. 

Building on \eqref{eq:intervention-via-mixture}, the mean outcome under a specific treatment $a'$ evaluates to:
\begin{align}
\mathbb{E}^{\mathrm{do}(A=a')}[Y]
&= \sum_{y\in\mathcal Y} y\,p^{\mathrm{do}(A=a')}(y) \nonumber \\
&= \sum_{\substack{y\in\mathcal Y\\ u\in\mathcal U}} y\,p(u)\,p(y\mid a',u).
\label{eq:ate-general}
\end{align}

Assuming a binary treatment $A\in\{0,1\}$, we can then define the \emph{average treatment effect (ATE)} as the contrast between the two interventional means:
\begin{align}
\mathrm{ATE} = \mathbb{E}^{\mathrm{do}(A=1)}[Y] - \mathbb{E}^{\mathrm{do}(A=0)}[Y].
\label{eq:ate}
\end{align}

At the subpopulation level, we can similarly define the conditional expected outcome for a given latent confounder $u$ under assignment $a'$. Relying on the invariance of the conditional distribution, we get
\begin{align}
\mathbb{E}^{\mathrm{do}(A=a')}[Y \mid U=u] &= \sum_{y\in\mathcal Y} y \, p^{\mathrm{do}(A=a')}(y \mid u) \nonumber \\
&= \sum_{y\in\mathcal Y} y \, p(y \mid a', u).
\label{eq:conditional-outcome}
\end{align}

Finally, the \emph{conditional average treatment effect} (CATE) is defined as the contrast between these conditional interventional means for a specific subpopulation:
\begin{align}
\mathrm{CATE}(u) = \mathbb{E}^{\mathrm{do}(A=1)}[Y \mid U=u] - \mathbb{E}^{\mathrm{do}(A=0)}[Y \mid U=u].
\label{eq:cate}
\end{align}

From \eqref{eq:intervention-via-mixture}, it becomes clear that if we can successfully recover $p(u)$ and the component conditional distributions $p(y\mid a,u)$ for all $a$, we can compute causal estimands directly. In fact, this form suggests a \emph{mixture learning perspective} for recovering the interventional distribution, the framework we will formalize later in the article.

\noindent {\bf The case of observed confounding.} \textcolor{black}{Before diving into the setup} with {\em unobserved} confounding, it is worth
\textcolor{black}{recalling the setting in which the confounder is} {\em observed}. 
In causal inference with observed confounding, estimation typically proceeds through one of two primary pathways:

\begin{itemize}[leftmargin=*]
    \item \emph{Modeling the outcome mechanism:} If we focus on how outcomes are generated, we aim to fit the conditional outcome model $p(y\mid a,u)$. By learning the conditional probability of observing an outcome given a treatment and an observed confounder $u$, we can identify the outcome mechanism.
    \item \emph{Modeling the treatment-assignment mechanism:} If we focus on how treatments are allocated, we aim to fit the conditional assignment model $p(a\mid u)$. By learning the conditional probability of receiving a treatment given an observed confounder $u$, we can identify the assignment mechanism.
\end{itemize}

If the treatment-assignment mechanism is known to be fully randomized across the population, the setting is referred to as a \emph{randomized controlled trial} (RCT). In this ideal scenario, the observational conditional outcome distribution $p(y\mid a)$ coincides with the interventional outcome distribution $p^{\mathrm{do}(A=a)}(y)$. However, in observational settings where \textcolor{black}{treatment allocation depends on the confounder}, this assignment-based approach falls under the framework of \emph{inverse propensity weighting} (IPW) \cite{ding2024first, pearl2009causal}. It is instructive to outline how IPW recovers interventional means by reweighting the observed outcomes based on a learned propensity model, rather than explicitly modeling the outcome mechanism.

\noindent \emph{Propensity scores and IPW.} We define the \emph{propensity score} for treatment level $a'$ as
\begin{align}
e_{a'}(u)=p(a'\mid u).
\label{eq:def-propensity}
\end{align}
This is the probability that a specific individual receives treatment based on their \textcolor{black}{characteristics determined by the confounder}. We assume positivity, i.e.,
$e_{a'}(u)>0$ for all $u\in\mathcal U$ with $p(u)>0$. Under the intervention $\mathrm{do}(A=a')$, from \eqref{eq:ate-general}, we can simplify $\mathbb{E}^{\mathrm{do}(A=a')}[Y]$ per \eqref{eq:def-propensity} as follows:
\begin{align}
\mathbb{E}^{\mathrm{do}(A=a')}[Y]
&=
\sum_{\substack{y\in\mathcal Y\\ u\in\mathcal U}} y\,p(u)\,p(y\mid a',u) \nonumber\\
&=  \sum_{u\in\mathcal U} p(u) \sum_{y\in\mathcal Y} y\,p(y\mid a',u) \nonumber\\
&=  \sum_{u\in\mathcal U} p(u) \frac{1}{p(a' \mid u)}
\sum_{y\in\mathcal Y} y\,p(y\mid a',u) p(a'\mid u) \nonumber\\
&=  \sum_{u\in\mathcal U} p(u) \frac{1}{p(a' \mid u)}
\sum_{y\in\mathcal Y} y\,p(y,a'\mid u)  \nonumber\\
&=  \sum_{u\in\mathcal U} p(u) \frac{1}{p(a' \mid u)}
\sum_{\substack{y\in\mathcal Y\\ a\in\mathcal A}} \mathbf{1}\{a=a'\}y\,p(y,a\mid u) \nonumber\\
&=  \sum_{u\in\mathcal U} p(u) \frac{1}{e_{a'}(u)}
\mathbb{E}[\mathbf{1}\{A=a'\}Y\mid U=u] \nonumber\\
&=
\mathbb{E}\left[
\frac{\mathbf{1}\{A=a'\}Y}{e_{a'}(U)}
\right].
\label{eq:ipw-final}
\end{align}

In a fully observed setting, suppose we had access to
$n$ independent samples $\{(u^{(k)},a^{(k)},y^{(k)})\}_{k=1}^n$
generated according to \eqref{eq:sem}. Let
\begin{align}
\mu(a') := \mathbb{E}^{\mathrm{do}(A=a')}[Y]
\end{align}
denote the interventional mean outcome under treatment $a'$. In general, the propensity score is also unknown and must be estimated from data using a method such as logistic regression. Given an estimate $\widehat e_{a'}(u)$ of $e_{a'}(u)$, the
plug-in IPW estimator is
\begin{align}
\widehat{\mu}_{\mathrm{IPW}}(a')
=
\frac{1}{n}\sum_{k=1}^{n}
\frac{\mathbf{1}\{a^{(k)}=a'\}y^{(k)}}{\widehat e_{a'}(u^{(k)})}.
\label{eq:ipw-empirical-plugin}
\end{align}

Consequently, if $U$ were observed, one could estimate the interventional mean under treatment $a'$ by reweighting each observed sample with $A=a'$ by the inverse probability of receiving that treatment within its type. However, in the setting of interest, $U$ is unobserved. Therefore, the sample-level weights $e_{a'}(u^{(k)})$ are unavailable.  The mixture-learning perspective discussed next seeks to recover the latent mixing distribution $p(u)$ and the component mechanisms $p(a\mid u)$ and $p(y\mid a,u)$ from the observed samples $(a^{(k)}, y^{(k)})$. Such recovery requires identifiability assumptions ensuring that the latent mixing distribution and component mechanisms are uniquely determined by the observational law.

\section*{Mixture learning}

\medskip
\noindent \textbf{\textcolor{black}{Mixtures.}} Let $U$ be a latent variable and $X$ be an observed variable, respectively supported on $\mathcal{U}$ and $\mathcal{X}$. The density of the observed random variable $X$, denoted $p(x)$, takes the form of a \emph{mixture} if:
\begin{align}
p(x) = \int_{u \in \mathcal{U}} p(u)\,p(x \mid u)\,du, \label{eq:general-mixture}
\end{align}
where $p(u)$, the density over the latent variable, is called the \emph{mixing} distribution \cite{fruhwirth2019handbook}. Given $u$, the conditional density of the observed random variable $X$, $p(x \mid u)$, is referred to as the \emph{component} distribution. A classical example dates back to Pearson's 1894 analysis where a population of crabs was modeled as a mixture of two normal distributions, suggesting the presence of latent subpopulations \cite{pearson1894iii, EM-royal}. 
The perspective of mixture learning poses the question: 
\begin{center}
    \emph{\textcolor{black}{Given} $n \geq 1$ independent and identically distributed (i.i.d.) samples $x^{(1)}, \dots, x^{(n)}$ drawn from the distribution $p(x)$ in \eqref{eq:general-mixture}, can we recover the underlying mixing distribution $p(u)$ and the component distributions $p(x \mid u)$?} 
\end{center}
While $\mathcal{U}$ may in general be continuous, we focus on the case where $\mathcal{U}$ is finite: $\mathcal{U} = \{u_1, u_2, \dots, u_M\}$ for $M \geq 1$. We can define the \emph{finite} mixing distribution over $\mathcal{U}$ with weights $\{w_1, w_2, \dots, w_M\}$ using Dirac delta functions as:
\begin{align}
p(u) = \sum_{m=1}^M w_m \delta(u - u_m). \label{eq:finite-mixing-density}
\end{align}
Plugging \eqref{eq:finite-mixing-density} into \eqref{eq:general-mixture}, we get the finite mixture form:
\begin{align}
p(x) = \sum_{m=1}^M w_m\,p(x \mid u_m). \label{eq:finite-mixture}
\end{align}
Of particular interest is the case when $p(x \mid u_m)$ takes a parametric form. Suppose there exists a deterministic, injective map $b: \mathcal{U} \to \Theta$, so that each $u_m$ is mapped to a unique $\theta_m$ for $m \in [M]$. \textcolor{black}{This induces a distribution} $P_\Theta(\cdot)$ over $\{\theta_1, \dots, \theta_M\} \subset \Theta$. \textcolor{black}{Define a \emph{parametric} mixture as}
\begin{align}
p(x) = \sum_{m=1}^M w_m\,p(x \mid \theta_m). \label{eq:parametric-mixture}
\end{align}
Suppose the parametric family $p(x \mid \theta)$ is known to us. The mixture learning question for finite parametric mixtures then entails recovering the mixing weights $\{w_m\}_{m \in [M]}$ and the component parameters $\{\theta_m\}_{m \in [M]}$ from $n \geq 1$ i.i.d. samples $x^{(1)}, \dots, x^{(n)}$ drawn from $p(x)$.

\medskip
\noindent \textbf{Identification and estimation for mixtures.} Mixture learning entails two fundamental questions: 
\begin{itemize}[leftmargin=*]
    \item \emph{Identification} asks whether the underlying mixture parameters can be uniquely recovered (\textcolor{black}{up to permutation} of labels) from the true, population-level observational distribution $p(x)$. If the mixture is non-identifiable, multiple distinct sets of mixing weights and component distributions could produce the exact same data distribution, making unique recovery impossible without further structural assumptions, such as linear independence among the component distributions.
    \item \emph{Estimation} deals with the \textcolor{black}{finite-sample challenge of recovering} parameters. Assuming the mixture is identifiable, we must computationally recover the parameters from the empirical samples $x^{(1)}, \dots, x^{(n)}$ drawn i.i.d. per $p(x)$. This involves navigating complex and often non-convex optimization landscapes.
\end{itemize}

Note that there are two distinct notions of identifiability. 
The first is \emph{causal identifiability}: can the target interventional quantity be expressed as a functional of the observational distribution under the structural assumptions encoded by the causal model? 
The second is \emph{mixture identifiability}: can the latent mixture weights and component distributions be uniquely recovered, at least up to label permutation, from that observational distribution? We refer the reader to \cite{pearl2009causal} and \cite{teicher1963identifiability}, respectively, for a comprehensive treatment of these two notions. Here, we shall assume that the given mixtures are identifiable.

\medskip\noindent\textbf{Approaches to mixture learning.}
Next, we present a brief overview of approaches to mixture learning:

\begin{itemize}[leftmargin=*]
    \item \emph{Empirical estimator:} Computes \textcolor{black}{a per-sample maximum likelihood estimate} to serve as a proxy for the latent variable, approximating the mixing distribution via the empirical distribution of these point estimates. While computationally simpler, its recovery accuracy is limited by localized sampling noise and \textcolor{black}{per-sample estimation error in the MLE}.

    \item \emph{Non-parametric MLE (NPMLE):} Discretizes the continuous support of the latent variable into finite bins to pool observational data. Geometrically, it performs a projection of the observed data frequencies onto the convex hull of the component distributions \cite{icml-vinayak19a}. It requires the computation of normalization constants typical of likelihood-based methods.

    \item \emph{Moment-based estimator:} Computes empirical estimates of moments of the underlying mixing distribution directly from observed outcomes. It then recovers the parameters by finding a probability measure, typically via a linear program, that closely matches these estimated moments \cite{tian2017learning}. 

    \item \emph{Expectation-Maximization (EM):} An iterative algorithm that alternates between computing the expected latent variables \textcolor{black}{given the current parameters} (the E-step) and updating the mixture parameters to maximize the expected complete-data log-likelihood (the M-step) \cite{balakrishnan2017statistical}. Geometrically, it operates by iteratively constructing and maximizing \textcolor{black}{a local {concave lower bound} on} the log-likelihood.
\end{itemize}

\textbf{Example: Mixtures of \textcolor{black}{i.i.d. Bernoulli Sequences}.}
To instantiate the mixture learning methods above, we consider learning mixtures of sequences of independent and identically distributed (i.i.d.) Bernoulli random variables. Formally, suppose we have $n$ i.i.d. samples, indexed by $k=1,\dots,n$, generated according to the following generative model:
\begin{tcolorbox}[
    colback=white,
    colframe=black,
    boxrule=0.6pt,
    arc=0pt,
    left=1pt,
    right=1pt,
    top=0pt,
    bottom=0pt,
    boxsep=1pt,
    before skip=8pt,
    after skip=8pt
]
\vspace{-0.5\baselineskip}
\begin{align}
    U &\sim \mathcal{P}_U, ~ U \in (0,1)
    \nonumber,\\
    X \mid U = u 
    &\sim \left\{\mathrm{Ber}(u)\right\}_{s=1}^t .
    \label{eq:toy-mix}
\end{align}
\vspace{-0.8\baselineskip}
\end{tcolorbox}
The primary objective is to learn the true underlying mixing distribution, $\mathcal{P}$, either by restricting it to a known, finite-dimensional parametric family or by non-parametrically estimating it as an unrestricted probability measure on $(0,1)$. {\color{black}We assume that $t$ is sufficiently large relative to $M$
so that the finite Bernoulli mixture is identifiable up to
label permutation. For an unrestricted mixing distribution,
identifiability is not guaranteed for finite $t$.} 

A simple empirical estimator computes the per-sequence maximum likelihood estimate (MLE) $\overline{X}^{(k)} = \frac{1}{t}\sum_{s=1}^t X_s^{(k)}$ as a plug-in estimate of $u^{(k)}$, and approximates the mixing distribution by the empirical CDF of these estimates. While computationally simple, its recovery accuracy is limited by within-sequence sampling noise and finite-sample error in the empirical distribution, achieving $\mathcal{O}_p\!\left(\frac{1}{\sqrt{n}}\right)+\mathcal{O}_p\!\left(\frac{1}{\sqrt{t}}\right)$ in   Wasserstein-1 ($W_1$) distance, also known as earth mover's distance \cite{icml-vinayak19a}.

Let $S^{(k)} = \sum_{s=1}^t X_s^{(k)}$. Since the observations are conditionally i.i.d. Bernoulli, $S^{(k)}$ is a sufficient statistic, and the model can equivalently be viewed as a mixture of Binomial random variables. For a candidate mixing distribution $Q$ on $(0,1)$, the per-sample likelihood of observing $S^{(k)}=s^{(k)}$ is
\begin{align}
    p_Q\!\left(s^{(k)}\right)
    =
    \int_0^1
    {t \choose s^{(k)}}
    u^{s^{(k)}}(1-u)^{t-s^{(k)}}
    dQ(u).
    \label{eq:binomial-mixture-marginal}
\end{align}
The nonparametric maximum likelihood estimator is then
\begin{align}
    \widehat{\mathcal{P}}_U
    \in
    \operatorname*{arg\,max}_{Q\in\mathcal{D}}
    \sum_{k=1}^n
    \log
    \left(
    \int_0^1
    {t \choose s^{(k)}}
    u^{s^{(k)}}(1-u)^{t-s^{(k)}}
    dQ(u)
    \right),
    \label{eq:npmle-binomial-mixture}
\end{align}
where $\mathcal{D}$ denotes the class of probability measures on $(0,1)$.  By effectively pooling information across the entire dataset, NPMLE significantly improves the $W_1$ error rate to $\mathcal{O}_p(1/t)$ when $t = \mathcal{O}(\log n)$, and $\mathcal{O}_p(1/\sqrt{t \log n})$ when $t = \Omega(\log n)$; see \cite{icml-vinayak19a} for precise scaling. \textcolor{black}{The moment-based estimator} matches the NPMLE when $t = \mathcal{O}(\log n)$; however, it fails when $t = \Omega(\log n)$ \cite{tian2017learning}. 

Fundamentally, NPMLE optimizes over all probability measures on the latent space. To render this infinite-dimensional problem computationally tractable, the latent space is discretized into a finite grid, reducing the task to fitting weights over these discrete points. However, this grid-based optimization scales poorly with the observation dimension, becoming computationally prohibitive in high-dimensional, dependent models burdened by intractable normalizing constants.  Furthermore, while Expectation-Maximization (EM) offers a workaround, it remains highly sensitive to initialization due to the non-convex likelihood landscape and is prone to getting trapped in local optima. The next section \textcolor{black}{introduces the exponential family as an extension} of the i.i.d. Bernoulli setting to allow for dependencies while retaining computational tractability of parameter learning.


\smallskip
\noindent\textbf{\textcolor{black}{Exponential families.}}
For modeling intricate dependencies in outcome trajectories, we introduce \textcolor{black}{the \emph{exponential family}, which} spans a broad class of parametric distributions \cite{wainwright2008graphical}.  Fix $D \in \mathbb{N}$ and let $\phi(x) = (\phi_1(x),\dots,\phi_D(x)) \in \mathbb{R}^D$ denote a vector of \emph{sufficient statistics}, where $x \in \mathcal{X}$. Then an exponential-family distribution can be written as
\begin{align}
p(x \mid \theta)
&\propto
\exp\!\left\{\left\langle \theta, \phi(x)\right\rangle\right\},
\label{eq:expfam-component}
\end{align}
where $\theta$ is the \emph{natural} parameter and
\begin{align}
A(\theta)
=
\log \int \exp\!\left\{\langle \phi(x),\theta\rangle\right\}\,dx,
\end{align}
is the \emph{log-partition}, or cumulant function. Standard maximum-likelihood \textcolor{black}{estimation for exponential families} is often computationally prohibitive in high dimensions due to intractable partition functions. Recent advances in \emph{M-estimation} \cite{shah2023counterfactualinferenceunobservedconfounding,shah2023computationally,vuffray2020} provide computationally efficient parameter \textcolor{black}{estimation for exponential families} under regularity and bounded domain assumptions. 
The key premise underlying these methods is a per-sample M-loss that bypasses the normalization required for traditional 
MLE while yielding a consistent and asymptotically normal estimator; see \cite{shah2023computationally} for precise 
guarantees.

We note that the Bernoulli distribution arises as a special case of the exponential family, with natural parameter given by the log-odds of success. Consider sequences of length $t$ of independent Bernoulli random variables \textcolor{black}{with the same underlying bias parameter} \eqref{eq:toy-mix}. Given the latent bias $u \in (0,1)$, \textcolor{black}{the conditional distribution of $x \in \{0,1\}^{t}$ over the $t$ time steps} is:
\begin{align*}
p\big(x \mid u \big) &= \exp\Biggl\{\,
\log\Biggl(
\prod_{s=1}^{t}u^{x_s}\,(1 - u)^{1 - x_s} 
\Biggr)
\Biggr\}\\
&= \exp\left\{ \sum_{s=1}^{t} x_s \log\left(\frac{u}{1 - u}\right) + t \log(1 - u) \right\} \\
&\propto \exp\left\{ \left\langle x, \theta(u) \right\rangle \right\}, \text{~where~} \theta(\cdot) = \log\left(\frac{\cdot}{1 - \cdot}\right) {\mathbf{1}}_{t \times 1}.
\end{align*}
\subsubsection*{From independent draws to pairwise dependencies}
While the i.i.d.\ Bernoulli sequence assumes each observation is independent given the latent bias, \textcolor{black}{real-world data often exhibit} local dependencies. For instance, in a temporal setting, the chance of a ``success'' today might directly impact the probability of a ``success'' tomorrow, such as an infection transmitting across consecutive days. Undirected graphical models, also known as \emph{Markov random fields} (MRFs), provide a natural framework for modeling such dependencies by explicitly modeling {\em sparsity} in model parameters to retain tractability. \textcolor{black}{Such models are effectively instances
of the exponential family}, cf. \cite{wainwright2008graphical}.

Of particular interest are \emph{pairwise} MRFs, which can capture pairwise interactions in the trajectory. Let the indices $s, j \in \{1, \dots, t\}$ represent the discrete steps in our sequence. We can extend the independent Bernoulli model by introducing the node-specific parameter vector $\theta(u) \in \mathbb{R}^t$ \textcolor{black}{dependent on a latent variable drawn from} $\mathcal{P}_U$ and a symmetric interaction matrix $\Lambda \in \mathbb{R}^{t \times t}$. The non-zero entries of $\Lambda$ dictate which steps interact (for example, adjacent time steps in a trajectory). The conditional distribution of the sequence $x \in \{0, 1\}^t$ given the latent variable $u$ takes the form:
\begin{align}
p(x \mid \theta(u), \Lambda) = \exp\left\{ \langle \theta(u), x \rangle + x^\top \Lambda x - A\bigl(\theta(u),\Lambda\bigr)\right\}, \label{eq:pairwisemrf}
\end{align}
where $A\bigl(\theta(u),\Lambda\bigr)$ is the log partition function. The linear term $  \langle \theta(u), x \rangle$ in \eqref{eq:pairwisemrf} captures \textcolor{black}{the node-specific biases, or `external fields', which} are analogous to the log-odds $\log\left(\frac{u}{1-u}\right)$ in the independent case. The quadratic term $x^\top \Lambda x $ captures the pairwise interactions governed by \textcolor{black}{the matrix $\Lambda$, with $\Lambda=0$} in the independent case \eqref{eq:toy-mix}.  The density \eqref{eq:pairwisemrf} generalizes \eqref{eq:toy-mix} as follows:
\begin{tcolorbox}[
    colback=white,
    colframe=black,
    boxrule=0.6pt,
    arc=0pt,
    left=1pt,
    right=1pt,
    top=0pt,
    bottom=0pt,
    boxsep=1pt,
    before skip=8pt,
    after skip=8pt
]
\vspace{-0.5\baselineskip}
\begin{align}
    U &\sim \mathcal{P}_U,
  \nonumber\\
    X \mid U = u
    &\sim p\bigl(\cdot \mid \theta(u), \Lambda\bigr).
    \label{eq:pairwise-mrf-generative}
\end{align}
\vspace{-0.8\baselineskip}
\end{tcolorbox}
In \eqref{eq:pairwise-mrf-generative}, the state space $\mathcal{X}$ may be discrete or continuous \cite{wainwright2008graphical,yang-ravikumar}, where $p(\cdot \mid \theta(u), \Lambda)$ is accordingly interpreted as either a probability mass function or a probability density function. When $\mathcal{P}_U$ takes the finite-mixture form in \eqref{eq:finite-mixing-density}, the log likelihood of $n$ i.i.d. samples $\{X^{(k)}\}_{k\in[n]}$ generated by $U^{(k)}\sim \mathcal{P}_U$ and
$X^{(k)}\mid U^{(k)}=u_m \sim p(\cdot\mid \theta(u_m),\Lambda)$ \eqref{eq:pairwisemrf} is
\begin{align}
    \frac{1}{n}\sum_{k=1}^{n}
    \log
    \left(
    \sum_{m=1}^{M}
    w_m
    \exp\left\{
        \sum_{s=1}^{t} \theta_s(u_m) x^{(k)}_s
        +  \sum_{s=1}^{t} \sum_{s'=1}^{t}
        \Lambda_{ss'} x^{(k)}_s x^{(k)}_{s'}
        - A\bigl(\theta(u_m),\Lambda\bigr)
    \right\}
    \right).
    \label{eq:expfam-mixture-loglikelihood}
\end{align}
Note that a mixture of exponential family distributions does not generally belong to the exponential family itself; learning their mixtures presents subtle identification and estimation challenges. 
Recent works have adapted M-estimation \cite{shah2023computationally} to the setting of learning mixtures of high-dimensional exponential \textcolor{black}{families \cite{ss_isit_2026,shah2023counterfactualinferenceunobservedconfounding} by first estimating} parameters $\widehat{\Lambda}$, $\{\widehat{\theta}(u^{(k)})\}_{k\in[n]}$  by minimizing the convex loss
\begin{align}
 \frac{1}{n} \sum_{k=1}^{n} \sum_{s=1}^{t} \exp\Biggl\{ -\biggl({\theta}_{s}(u^{(k)}) + 2 \sum_{s' \in [t]\setminus\{s\}}{\Lambda}_{ss'}\, x^{(k)}_{s'}\biggr) \Bigl(x^{(k)}_s - \mathbb{E}_{\mathrm{Unif}_{\mathcal{X}}}[.]\Bigr) \Biggr\}, \label{eq:pairwise-m-loss}
\end{align}
where the expectation is with respect to the uniform random variable
supported on $\mathcal{X}$, with $\mathcal{X}$ being bounded,
\textcolor{black}{and the natural parameters are subject to
appropriate norm bounds}.
\textcolor{black}{Recent work \cite{sood2026learning} extends this
approach to jointly estimate shared interaction parameters and
component-specific linear parameters for structured mixtures
with a small number of components. Under suitable separation
and regularity conditions, this enables recovery of mixture
components and their latent labels.}


\section*{Mixture learning and causal inference with unobserved confounding} One way to see the connection between mixtures and causal estimands is by comparing the form of \eqref{eq:intervention-via-mixture} and \eqref{eq:general-mixture}. We see that causal inference with latent confounding naturally relates to a mixture distribution. Specifically, for each action $a'$, the target interventional distribution $p^{\mathrm{do}(A=a')}(y)$ is a mixture over latent confounder types, with mixing weights $p(u)$ and action-specific component densities $p(y\mid a',u)$. This will lead
to \textcolor{black}{causal estimands such as ATE and CATE} as discussed in \eqref{eq:ate-general} and \textcolor{black}{\eqref{eq:conditional-outcome}, respectively}. Therefore, we shall focus on establishing the following: 
\begin{center}
    {\em \textcolor{black}{How can mixture learning help learn the distribution of the unobserved confounder $U$ and the conditional distribution of
    the outcome $Y$ given the action $A$ and the confounder $U$?}}
\end{center}
\begin{tcolorbox}[
    colback=white,
    colframe=black!25,
    boxrule=0.4pt,
    arc=1pt,
    left=4pt,
    right=4pt,
    top=3pt,
    bottom=3pt,
    boxsep=0pt,
    before skip=7pt,
    after skip=7pt
]
{
\noindent{\bf Scope and key assumptions.} We are focusing on a rather generic setting with 
minimal assumptions. In effect, the purpose of the study is to \textcolor{black}{understand the impact of the intervention}
$A$ on outcome $Y$. All the other aspects of the causal mechanism are encoded through $U$, which 
may be potentially high-dimensional, and it is considered to be unobserved or latent. \textcolor{black}{This sets up
the mechanism} as described in \eqref{eq:sem}: $U\to A$, $U\to Y$, and $A\to Y$. While the setting 
is generic, to develop meaningful results, we \textcolor{black}{consider $U$ to be discrete with a known bound on
its cardinality}. This naturally \textcolor{black}{yields a connection to the setting of finite mixtures; see} \eqref{eq:finite-mixture}. To enable identifiability and estimation, depending upon the setting, appropriate 
assumptions are made. Indeed, making these as generic as possible is the primary intellectual quest here. 
}
\end{tcolorbox}
Under a parametric representation in which the latent confounder type is indexed by $\theta_m$, the interventional distribution for a specific action $A=a'$ can be written as
\begin{align}
p^{\mathrm{do}(A=a')}(y)
=
\sum_{m=1}^M w_m\,p(y\mid \theta_m,a').
\end{align}
\noindent Thus, for each action $a'\in\mathcal A$, the interventional distribution $p^{\mathrm{do}(A=a')}(y)$, and subsequently causal estimands such as the average treatment effect via \eqref{eq:ate}, can be derived once the shared latent population weights $w_m$ and the action-specific component mechanisms $p(y\mid \theta_m,a')$ have been learned. {For instance, in Fig.~\ref{fig:simpson-mixture}, learning the two latent diabetes types together with their corresponding dose-response relationships
$p(y\mid \theta_m,a')$ allows us to recover the true negative effect of
insulin that is obscured when the data are aggregated across the two types.}

Now we concretely \textcolor{black}{connect the mixture learning question to the causal inference question} from observational data.  \textcolor{black}{Recall our setting where, following the mechanism} in \eqref{eq:sem}, $n$ independent samples $(u^{(k)}, a^{(k)}, y^{(k)})$ are generated for $k \in [n]$. In our setting with unobserved confounding, we only observe $(a^{(k)}, y^{(k)})$ for $k \in [n]$. As is standard in many mixture-learning formulations, we assume that the model class, the number of latent components, and the relevant supports are specified. 

\noindent\textbf{Learning the confounder distribution.}
Depending on what we know about \textcolor{black}{the joint distribution of $(U,A,Y)$} and the dimensionality of $\mathcal{A}$ and $\mathcal{Y}$, we could consider the following mixture learning formulation to learn the distribution of latent confounders \textcolor{black}{(viewed as a mixing density)}.

\begin{itemize}[leftmargin=*]
    \item \emph{Learning from $x=a$:} If we use only the observed sequence of treatments $\{a^{(k)}\}_{k \in [n]}$, we can leverage the fact that the marginal law $p(a)$ satisfies
    \begin{align}
    p(a)=\sum_{u\in\mathcal U} p(u)\,p(a\mid u). \label{eq:learn-from-a}
    \end{align}
    Notice that the component distribution here, $p(a\mid u)$, is precisely the propensity score $e_a(u)$ defined in \eqref{eq:def-propensity}. Thus, if we can successfully identify and estimate this mixture model, we can recover both the distribution of confounders and the latent propensity scores. \textcolor{black}{Using the information thus learned to enable methods such as inverse propensity weighting (IPW) would additionally require assigning a latent confounder to each sample, i.e., identifying} $u^{(k)}$ for each $k$. \textcolor{black}{If treatment data are rich enough, such a method can work}. However, in many settings, treatment is low-dimensional (e.g. $0$ or $1$) and hence insufficient to identify both $p(u)$ and $p(a\mid u)$. \textcolor{black}{Such settings necessitate utilizing information beyond the action, which} we discuss next. 

    \item \emph{Learning from $x=y$:} If we use only the observed outcome sequence  $\{y^{(k)}\}_{k \in [n]}$ drawn per the marginal 
    \begin{align}
    p(y) = \sum_{u\in\mathcal U} p(u)\,p(y\mid u),
    \end{align}
    again we can see how recovering the latent confounder distribution maps to a mixture learning problem. While resolving this mixture captures latent heterogeneity, it is fundamentally insufficient for causal inference. \textcolor{black}{The learned component $p(y \mid u)$ inherently aggregates over all treatment assignments and cannot isolate} the action-specific mechanism $p(y\mid a',u)$ required to evaluate interventional estimands.
    
    \item \emph{Learning from $x=(a,y)$:} By treating the joint action-outcome pair as the observation, we have
    \begin{align}
        p(a,y) = \sum_{u\in\mathcal U} p(u)\,p(a,y\mid u).
    \end{align}
    Fitting this joint distribution from the action-outcome sequence
    $\{(a^{(k)},y^{(k)})\}_{k \in [n]}$ can provide a higher-dimensional signal,
    depending on the supports $\mathcal{A}$ and $\mathcal{Y}$. Under appropriate
    identifiability conditions, this approach can recover the latent mixing
    weights $p(u)$ along with the joint component distribution $p(a,y\mid u)$.
    From each recovered joint component, we can obtain the treatment-assignment
    mechanism by marginalizing over outcomes:
    \begin{align}
        p(a\mid u) = \sum_{y\in\mathcal Y} p(a,y\mid u).
    \end{align}
    Thus, under mixture identifiability and the corresponding positivity
    condition, the recovered weights $p(u)$ and conditional outcome mechanisms
    $p(y\mid a',u)$ can be combined to reconstruct the target interventional
    distribution $p^{\mathrm{do}(A=a')}(y)$.
 \end{itemize}
 Having established a direct correspondence between computing causal estimands and learning mixtures, we next 
 discuss computationally efficient methods for causal inference with unobserved confounding that rely on mixture learning approaches such as M-estimators, \textcolor{black}{latent factor models, and the method of moments}. En route, we instantiate these methods \textcolor{black}{as extensions of i.i.d. Bernoulli mixtures that allow arbitrary pairwise dependencies in outcome trajectories and time-varying bias parameters}.

\noindent \textbf{Bernoulli mixtures and panel data.}
Next, we extend the Bernoulli mixture setting to the causal context, as illustrated by the DAG in Figure~\ref{fig:simpson-mixture}(b), through the following generative mechanism. For $k \in [n]$, $(u^{(k)}, a^{(k)}, y^{(k)})$ are sampled as:
\begin{tcolorbox}[
    colback=white,
    colframe=black,
    boxrule=0.6pt,
    arc=0pt,
    left=1pt,
    right=1pt,
    top=0pt,
    bottom=0pt,
    boxsep=1pt,
    before skip=8pt,
    after skip=8pt
]
\vspace{-0.5\baselineskip}
\begin{align}
    U &\sim \mathcal{P}_U
    &&\text{(latent confounder)}, ~ U \in (0,1)  \nonumber,\\
    A \mid U = u
    &\sim \left\{\mathrm{Ber}\bigl(q(u)\bigr)\right\}_{s=1}^t
    &&\text{(action sequence)}, ~ A \in \{0,1\}^t \nonumber,\\
    Y \mid A=a, U = u
    &\sim \left\{\mathrm{Ber}\bigl(r(u,a_s)\bigr)\right\}_{s=1}^t
    &&\text{(outcome sequence)}, ~ Y \in \{0,1\}^t .
    \label{eq:confounded-sequence-model}
\end{align}
\vspace{-0.8\baselineskip}
\end{tcolorbox}
After sampling $u^{(k)}$ from $\mathcal{P}_U$, we generate the action/treatment sequence $a^{(k)}$ from a Bernoulli sequence parametrized by the confounder $u^{(k)}$. Compared to the vanilla Bernoulli sequence mixture in \eqref{eq:toy-mix}, the generative mechanism in \eqref{eq:confounded-sequence-model} allows each outcome $y^{(k)}_s$ to depend on both the latent confounder $u^{(k)}$ and the index-specific action $a^{(k)}_s$. In this context, a `panel' refers to a sequence of repeated measurements taken over time on multiple units/individuals. 
{\color{black}Throughout, we assume positivity: for every latent support point $u_m$,
$q(u_m)\in(0,1)$, $r(u_m,a)\in(0,1)$ for $a\in\{0,1\}$,
and the mixing weights $w_m>0$.
Further, assume $q(\cdot), r(\cdot,\cdot)$ are injective on the latent support
and $t$ is sufficiently large relative to $M$.}

We assume that the mixing distribution $\mathcal{P}_U$ is finite and parametric, as in \eqref{eq:finite-mixing-density}, and takes the form $\sum_{m=1}^M w_m \delta(u-u_m)$. To compute causal estimands such as the interventional distribution in \eqref{eq:intervention-via-mixture}, we can use a mixture-learning approach.  \textcolor{black}{Recall that to infer the interventional distribution} \eqref{eq:intervention-via-mixture}, we need to learn the confounder distribution $p(u)$, and for each action, the component $p(y\mid a,u)$. One pathway is to use MLE or EM to optimize for the joint likelihood\textcolor{black}{:} 
\begin{align}
 \frac{1}{n} \sum_{k=1}^n \log \left( \sum_{m=1}^M w_m \prod_{s=1}^t \Big[ q(u_m)^{a^{(k)}_s} \big(1-q(u_m)\big)^{1-a^{(k)}_s} r(u_m, a^{(k)}_s)^{y^{(k)}_s} \big(1-r(u_m, a^{(k)}_s)\big)^{1-y^{(k)}_s} \Big] \right).
\label{eq:joint-mle-objective}
\end{align}
\textcolor{black}{Here,} \eqref{eq:joint-mle-objective} is maximized over the parameters $\{w_m,q(u_m),r(u_m,0),r(u_m,1)\}_{m=1}^M$, subject to the mixing weights $\{w_m\}_{m=1}^M$ lying in the probability simplex, i.e., $w_m\geq 0$ and $\sum_{m=1}^M w_m=1$, and the Bernoulli parameters $q(u_m),r(u_m,a)$ lying in $(0,1)$ for $a\in\{0,1\}$.


Alternatively, we may first use the action panels $\{a^{(k)}_s\}_{s \in [t],\, k \in [n]}$ to learn a Bernoulli mixture with component biases $\{q(u_m)\}_{m=1}^M$ and corresponding mixing weights $\{w_m\}_{m=1}^M$. Since we assumed $q(\cdot)$ is injective on the latent support, these recovered biases identify the latent components up to label permutation. Given the learned $\{w_m,q(u_m)\}_{m=1}^M$, we then estimate the remaining outcome parameters $\{r(u_m,0),r(u_m,1)\}_{m=1}^M$ from the observed outcome panels and their associated action sequences.

With the mixing weights $\{w_m\}_{m=1}^M$ and the outcome mechanisms $\{r(u_m,0), r(u_m,1)\}_{m=1}^M$ learned, we can reconstruct the target interventional distribution. Applying \eqref{eq:intervention-via-mixture} to this panel setting, the probability of observing an outcome sequence $y \in \{0,1\}^t$ under a target intervention sequence $a' \in \{0,1\}^t$ is
\begin{align}
p^{\mathrm{do}(A=a')}(y)
=
\sum_{m=1}^M w_m
\prod_{s=1}^t
\Big[
r(u_m,a'_s)^{y_s}
\big(1-r(u_m,a'_s)\big)^{1-y_s}
\Big].
\label{eq:panel-interventional-distribution}
\end{align}
The setting in \eqref{eq:confounded-sequence-model} corresponds to the case where, for each individual sample, actions across time indices are assigned independently conditional on the latent confounder.  {\color{black}Under $q(u)\in(0,1)$ for all latent components $u$, both treatment
values have positive probability within each latent component.
Hence, for sufficiently long panels, the same individual unit may
contain observations under both treatment values, supporting
identification under appropriate conditions.}

\noindent \textbf{Latent factor models and synthetic interventions.}
Next, we consider a more realistic setting in which the action is \emph{coupled} across time indices, such that each unit may be observed under \emph{only one treatment}. Recalling Example 1 on the tobacco policy, we note that each state either permanently adopted a policy or maintained the status quo. Admittedly, estimating causal quantities is significantly more challenging in this regime because we lose within-unit treatment variation. Suppose we seek to answer the question: what would have happened if a treated state that imposed a specific anti-tobacco program had instead raised taxes? This requires evaluating a \emph{counterfactual} at the individual unit level under mutually exclusive treatment scenarios \footnote{It is crucial to distinguish unit-level counterfactuals from interventional quantities defined via the $do$-operator, as the latter characterize outcome distributions under an externally imposed treatment rather than targeting the unobserved potential outcome of a specific unit.}.


To resolve this, we turn to approaches that exploit repeated measurements over time under additional structural restrictions \cite{survey,ding2024first,did-abadie,agarwal2025model,agarwal2025synthetic}. \emph{Synthetic control} \cite{abadie2010synthetic} methods construct a counterfactual for a treated unit by finding a \emph{weighted} combination of untreated units that closely tracks the treated unit's trajectory prior to the intervention. By leveraging the temporal structure of the data, this approach effectively accounts for unobserved confounding, provided the latent confounding factors can be represented by a stable low-rank factor model. The \emph{synthetic interventions} \cite{agarwal2025synthetic} framework generalizes this to multiple treatments, effectively mapping the counterfactual estimation problem \textcolor{black}{to a variation of the standard \emph{tensor completion} problem -- see, for example,} \cite{ChenChiFanMa2020}
for the {\em standard} setting of tensor completion to compare and contrast the setting relevant to synthetic interventions.
\begin{figure}[t]
\centering
\includegraphics[width=0.99\linewidth]{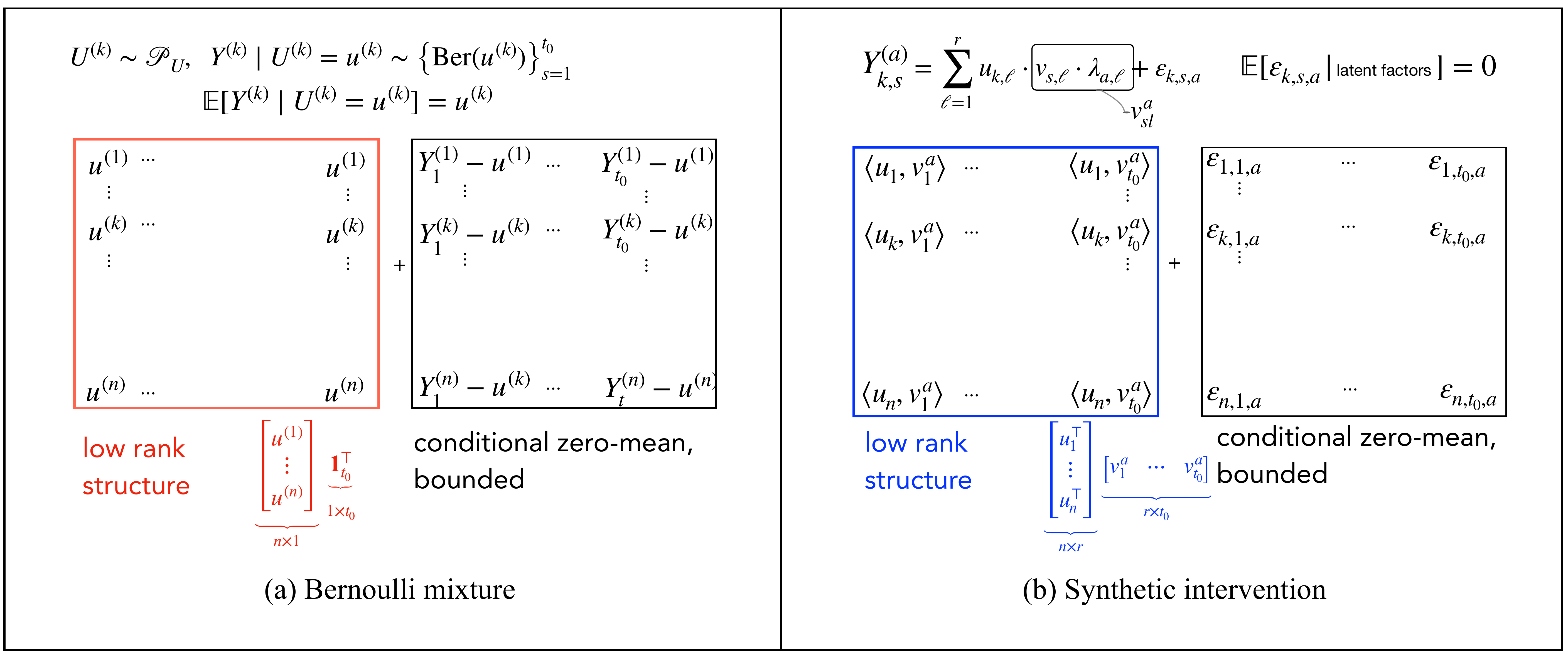}
\caption{From mixture learning to synthetic interventions: (a) A Bernoulli mixture \eqref{eq:toy-mix} induces a low-rank matrix structure because each unit has a latent row-level parameter. (b) Synthetic interventions extend this idea to a low-rank unit, time, treatment tensor.}
\label{fig:mix-to-SI}
\end{figure}
Before introducing the tensor-based framework, we revisit the Bernoulli mixture model in \eqref{eq:toy-mix} and view it as a matrix completion problem (Figure~\ref{fig:mix-to-SI}(a)). Let $Y\in\{0,1\}^{n\times t}$ denote the observation matrix. Conditional on the latent parameter $u^{(k)}$, the mean of row $k$ is constant across time and equals $u^{(k)}$. Thus, under the finite mixture model in \eqref{eq:finite-mixing-density}, the mean matrix has only $M$ distinct row types, inducing a low-rank clustered-row structure. \textcolor{black}{Truncated singular value decomposition (SVD)} \cite{chatterjee2015matrix} or 
\textcolor{black}{nearest-neighbor methods such as collaborative filtering} \cite{blindreg} can exploit this structure to denoise the observations and estimate 
the latent sequence-level parameters \cite{ChenChiFanMa2020}.

\begin{figure}[t]
\centering
\includegraphics[width=0.7\linewidth]{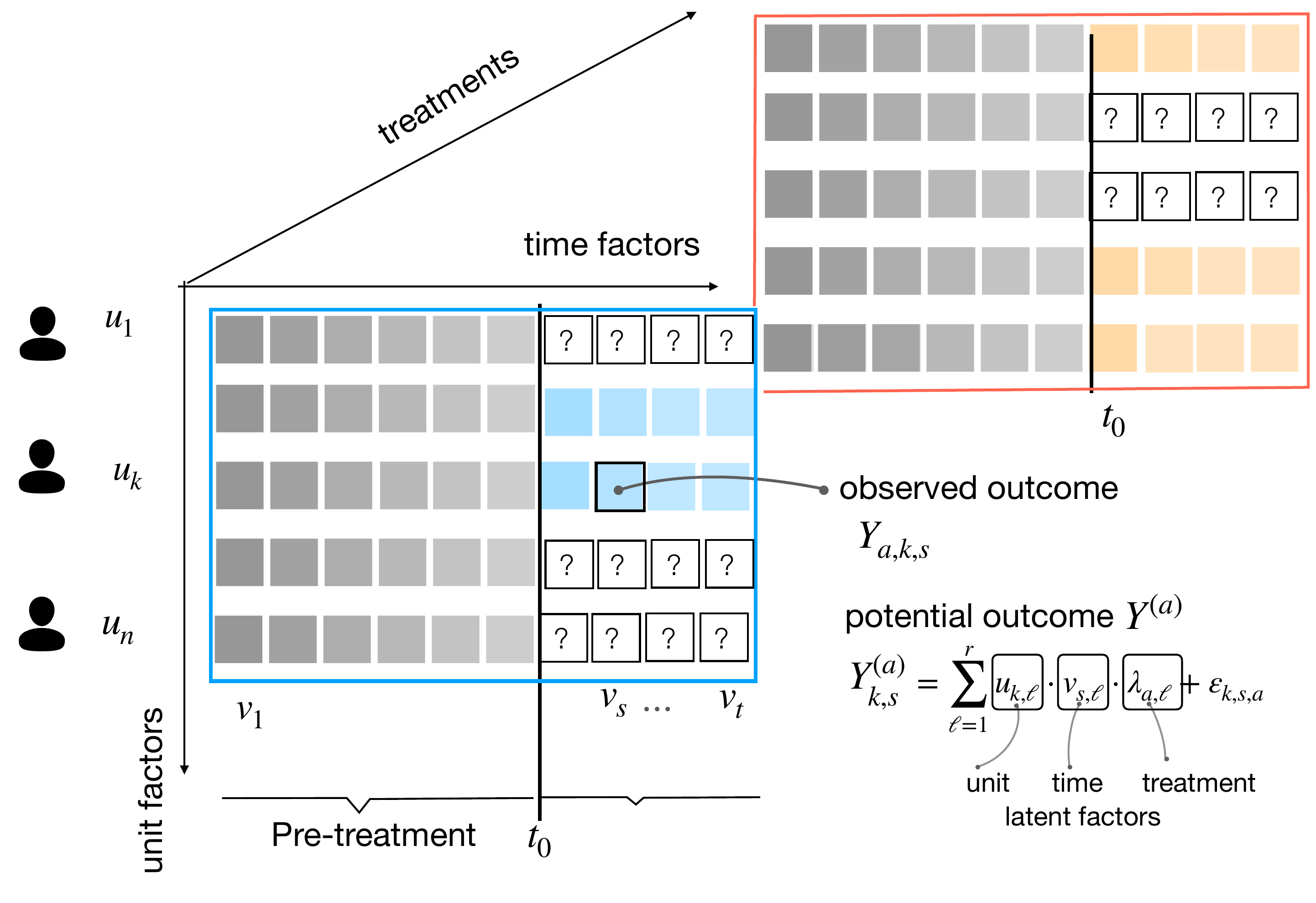}
\caption{Synthetic intervention view of unobserved confounding. Potential outcomes are modeled through unit, time, and treatment factors, inducing a low-rank tensor structure. Tensor completion uses the observed entries to estimate missing post-intervention counterfactual outcomes.} \label{fig:synthetic-interventions-latent-factor}
\end{figure}

Synthetic interventions (SI) extend this low-rank framework in two key ways: by incorporating time-level latent factors into the expected potential outcomes, and by introducing treatment as a third mode. The central premise is that expected potential outcomes across units, time, and treatments admit a low-rank tensor factorization, enabling counterfactual entries to be estimated by tensor completion (Figure~\ref{fig:mix-to-SI}(b)). For unit $k$ at time $s$, let $Y_{k,s}^{(a)}$ denote its potential outcome under treatment $a$. Both the SI and SC frameworks operate under the stable unit treatment value assumption (SUTVA), which in essence means that treatments are well-defined and each unit's potential outcome under a given treatment depends only on the treatment assigned to that unit, not on the treatments assigned to other units. Specifically, the SI framework models $Y_{k,s}^{(a)}$ for unit $k$, at time $s$, under treatment $a$ \textcolor{black}{through interactions among unit factors} $u_k$, time factors $v_s$, and treatment factors $\lambda_a$ (Figure~\ref{fig:synthetic-interventions-latent-factor}) as
\begin{align} 
Y_{k,s}^{(a)} = \sum_{\ell=1}^{r} u_{k,\ell} \cdot v_{s,\ell} \cdot \lambda_{a,\ell} + \varepsilon_{k,s,a}
\end{align} 
where $r$ is assumed to be much smaller than the number of units assigned to treatment $a$ and the number of pre-intervention time periods, during which outcomes are observed for all units, denoted by $t_0$.
After $t_0$, each unit is observed under only one specific treatment, leaving counterfactual outcomes under alternative treatments as missing entries (Figure~\ref{fig:synthetic-interventions-latent-factor}). The causal estimand of interest is the average expected potential outcome for a target unit $k$ under a specific treatment $a'$ during the post-treatment period $\{t_0+1,\cdots, t_f\}$, given by
\begin{align} 
\frac{1}{t_f - t_0} \sum_{s=t_0+1}^{t_f} \mathbb{E} \left[ Y_{k,s}^{(a')} \mid \{u_k\}, \lambda_{a'}, \{v_s\}_{s>t_0} \right].
\label{eq:si-causal}
\end{align} 
To estimate \eqref{eq:si-causal}, the SI framework relies on three key assumptions \cite{agarwal2025synthetic}. First, during the pre-treatment period, the observed outcomes match the potential outcomes, while in the post-treatment period, the outcome is recorded for only one assigned intervention per unit $k$ and is missing otherwise. Second, noise is mean-zero conditional on the observation pattern $\mathcal{D}$ and the latent factors. Finally, a \emph{latent span condition} assumes the target unit $k$ can be reconstructed as a \emph{weighted combination} of donor units assigned to the target treatment $a'$. Under these assumptions, the SI framework uses the fully observed pre-treatment entries to learn the underlying low-rank factors and projects these learned factors to impute the missing post-intervention counterfactual outcomes for the targeted units. Specifically, an estimator based on principal component regression (SI-PCR) is employed, yielding a statistically consistent estimate of the causal estimand \textcolor{black}{under additional subspace
and regularity conditions}; see \cite{agarwal2025model, agarwal2025synthetic} for precise guarantees.

While powerful, the tensor-based SI approach has a few key limitations. The framework primarily estimates conditional mean potential outcomes rather than the full distribution of counterfactual outcomes. It often relies on structured observation patterns, such as observing all units in the pre-treatment period. Finally, repeated measurements for the same unit are generally dependent across time, so statistical dependence across time indices must either be modeled explicitly or controlled through additional assumptions.

\noindent \textbf{Modeling dependencies in treatment-outcome panels via exponential families.}
Next, we model counterfactual distributions, rather than only counterfactual means as in the SI framework. We also allow arbitrary pairwise interactions among latent confounders, actions, and outcomes. To model dependencies in $(U,A,Y)$, we invoke the exponential-family form in \eqref{eq:expfam-component}.
Suppose $u \in \mathcal{U} \subseteq \mathbb{R}^{d_u}$, $a \in \mathcal{A} \subseteq \mathbb{R}^{d_a}$, and $y \in \mathcal{Y} \subseteq \mathbb{R}^{d_y}$, \textcolor{black}{with bounded domains}. Let $\phi$ denote the collection of linear parameters and $\Phi$ the collection of pairwise interaction parameters. We define the joint distribution
\begin{align}
f_{U,A,Y}(u,a,y; \phi, \Phi)
\propto
\exp\left\{
\begin{aligned}
& ~~~~y^\top \phi^y  
+ y^{\top} \Phi^{yy} y
+ 2 y^{\top} \Phi^{yu} u \\
&+ a^\top  \phi^a 
+ a^{\top} \Phi^{aa} a
+ 2 a^{\top} \Phi^{ay} y \\
&+ u^\top \phi^u  
+ u^{\top} \Phi^{uu} u
+ 2 a^{\top} \Phi^{au} u
\end{aligned}
\right\}.
\label{eq:expf-joint}
\end{align}

\begin{figure}[!t]
\centering
\includegraphics[width=0.79\linewidth]{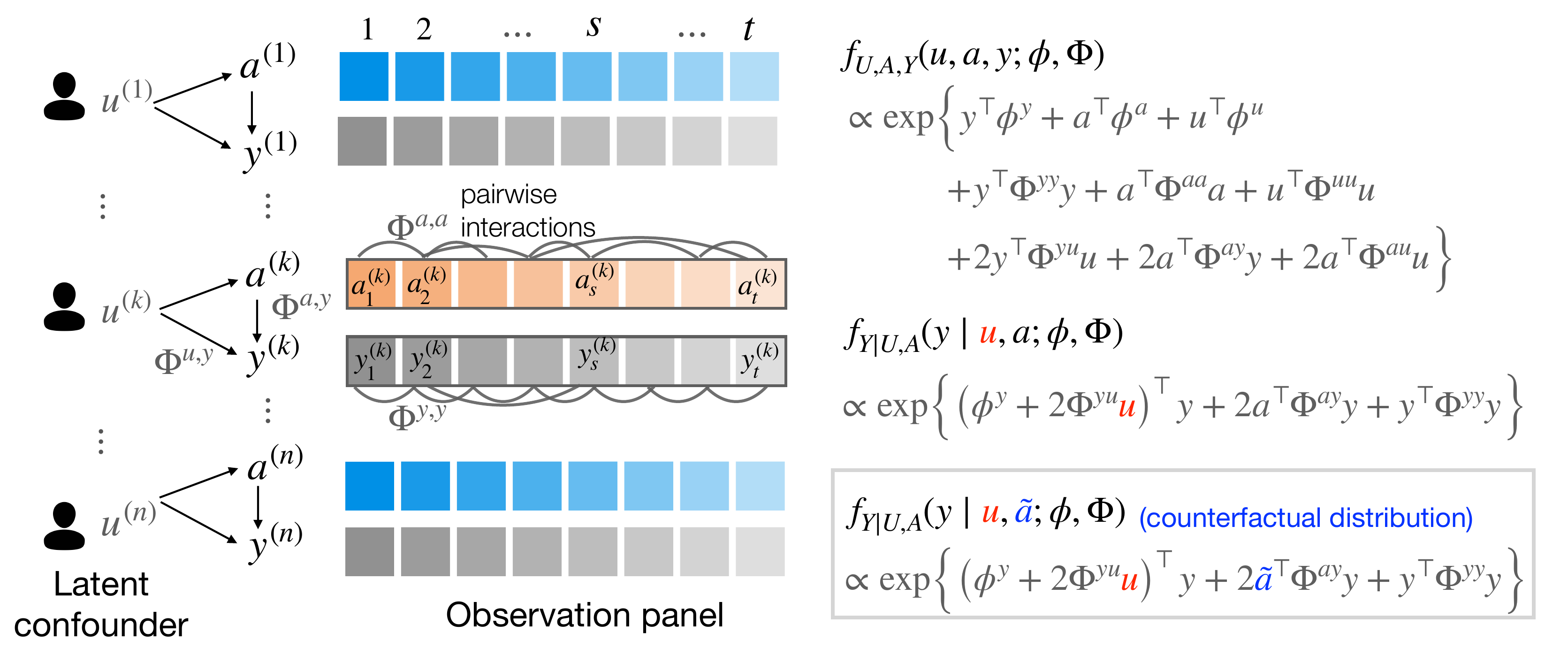}
\caption{Exponential-family model for treatment-outcome panels. Latent confounders induce dependencies among actions and outcomes, and the learned conditional distribution $f_{Y\mid U,A}$ supports counterfactual evaluation under alternative actions $\widetilde a$.}
\label{fig:method-exp-family-m-estimator}
\end{figure}

 Under the joint model \eqref{eq:expf-joint}, let $f_U(\cdot)$, $f_{A\mid U}(\cdot\mid\cdot)$, and $f_{Y\mid A,U}(\cdot\mid\cdot,\cdot)$ denote the induced marginal and conditional distributions. Following our canonical causal mechanism in Fig.~\ref{fig:simpson-mixture}(b), we generate a panel (Fig.~\ref{fig:method-exp-family-m-estimator}) by sampling for $k \in [n]$,
$U^{(k)}$ from $f_U(\cdot)$,
$A^{(k)} \mid U^{(k)} =u^{(k)}$ from $f_{A \mid U}(\cdot \mid u^{(k)})$,
and $Y^{(k)} \mid A^{(k)}=a^{(k)},U^{(k)}=u^{(k)}$ from
$f_{Y \mid A,U}(\cdot \mid a^{(k)},u^{(k)})$.


Given the observed panel $\{(a^{(k)},y^{(k)})\}_{k=1}^n$, our goal is to estimate counterfactual outcome distributions under alternative treatments. \textcolor{black}{Under the stable unit treatment value assumption} (SUTVA),  changing the treatment assigned to unit $k$ does not affect the outcomes of other units. Under the structural model \eqref{eq:sem}, setting the treatment of unit $k$ to $a$ leaves the outcome mechanism unchanged; hence, conditional on $U^{(k)}=u$, its counterfactual outcome under treatment $a$ has distribution $f_{Y\mid A,U}(\cdot\mid a,u)$. Thus, learning these conditional counterfactual distributions reduces to learning $f_{Y\mid A,U}(\cdot\mid a,u)$ for $u\in\mathcal{U}$ and $a\in\mathcal{A}$.

For the exponential-family model \eqref{eq:expf-joint}, conditional distributions obtained by conditioning on subsets of variables retain a similar exponential-family form. From \eqref{eq:expf-joint}, the conditional distribution 
\begin{align}
f_{Y \mid A,U}(y \mid a,u; \phi, \Phi)
\propto
\exp\left\{
(\phi^y + 2 \Phi^{yu} u)^{\top} y
+ 2 a^{\top} \Phi^{ay} y
+ y^{\top} \Phi^{yy} y
\right\}.
\label{eq:expf-conditional}
\end{align}
To learn \eqref{eq:expf-conditional}, we can leverage the literature on mixture learning for exponential families \cite{ss_isit_2026, sood2026learning, fruhwirth2019handbook}. Let
\begin{align}
x =
\begin{bmatrix}
a \\
y
\end{bmatrix}
\in \mathbb{R}^{d_a+d_y}
\end{align}
denote the concatenated action-outcome vector. 
Conditioning \eqref{eq:expf-joint} on $u$ yields
\begin{align}
f_{X \mid U}(x \mid u; \phi, \Phi)
\propto
\exp\left\{
x^{\top}
\begin{bmatrix}
\phi^a + 2\Phi^{au}u \\
\phi^y + 2\Phi^{yu}u
\end{bmatrix}
+
x^{\top}
\begin{bmatrix}
\Phi^{aa} & \Phi^{ay} \\
\Phi^{ya} & \Phi^{yy}
\end{bmatrix}
x
\right\}.
\label{eq:expf-x}
\end{align}
Comparing \eqref{eq:expf-x} with \eqref{eq:pairwisemrf}, we see that \eqref{eq:expf-x} is obtained by taking
\begin{align}
\theta(u)
=
\begin{bmatrix}
\phi^a + 2\Phi^{au}u \\
\phi^y + 2\Phi^{yu}u
\end{bmatrix},
\qquad
\Lambda
=
\begin{bmatrix}
\Phi^{aa} & \Phi^{ay} \\
\Phi^{ya} & \Phi^{yy}
\end{bmatrix}.
\end{align}
Thus, estimating counterfactual distributions \textcolor{black}{has a direct correspondence to mixture learning} for exponential families \cite{ss_isit_2026}, as discussed earlier. This provides another example of the synergy between causal inference with unobserved confounding and mixture learning. Recent work \cite{sood2026learning} further connects latent
heterogeneity to the calibration of physical measurements
in the presence of hardware variability.

Open problems include extending this framework to exponential families with unbounded domains and to higher-order sufficient statistics beyond quadratic interactions, especially when such statistics are confounder-dependent.

\begin{figure}[t]
    \centering
    \includegraphics[width=0.94\linewidth]{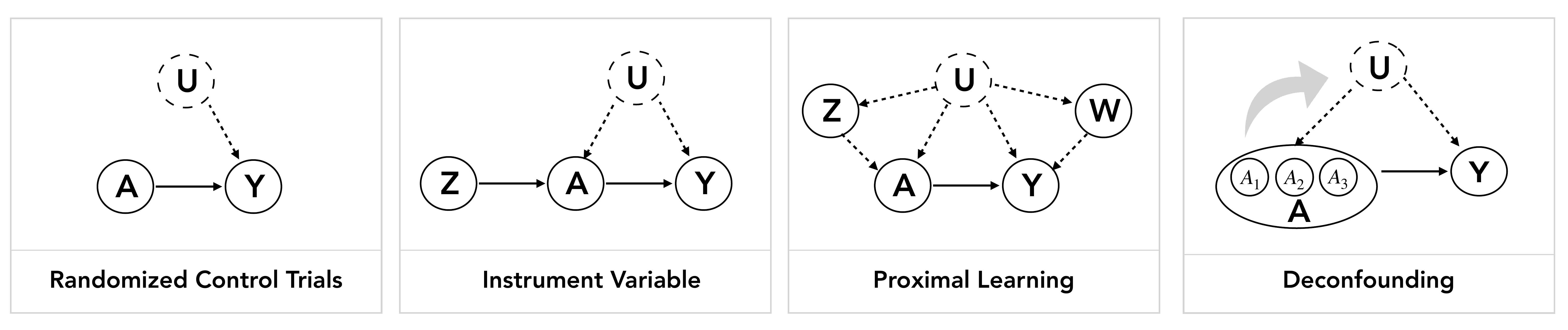}
    \caption{The canonical causal DAG (Fig.~\ref{fig:simpson-mixture}) contains an unobserved confounder $U$ affecting both treatment $A$ and outcome $Y$. \textcolor{black}{Randomized controlled trials} externally assign $A$ to break its dependence on $U$; instrumental variables use \textcolor{black}{an external source of variation $Z$} affecting $Y$ only through $A$; proximal learning uses observed proxies $Z$ and $W$ carrying information about $U$; and deconfounders leverage the multiplicity of causes to infer a latent substitute confounder by modeling the joint treatment distribution.}
    \label{fig:other-methods}
\end{figure}

\noindent \textbf{Other approaches to unobserved confounding.} The challenge of unobserved confounding can be overcome in multiple ways: from randomized experimentation to natural experimentation, also known as \emph{instrumental variables} (IV), pioneered \textcolor{black}{in the early econometrics literature} to isolate exogenous \textcolor{black}{variation in the intervention} \cite{iv}. While randomized experiments require controlled environments, IV applies to observational data, the primary interest of this article. For observational data, IV leverages an external variable $Z$ to isolate exogenous variation in \textcolor{black}{the treatment $A$ (Fig.~}\textcolor{black}{\ref{fig:other-methods}). To successfully} bypass the latent confounder $U$ and consistently estimate causal effects, a valid instrument must causally influence $A$, affect the outcome $Y$ strictly through its effect on $A$, and share no unobserved confounders with $Y$. 

In practice, however, finding valid and strong instruments remains a significant bottleneck. When external instruments are unavailable, the \emph{deconfounder} method \cite{blei-deconfounder} offers an alternative by leveraging the multiplicity of causes. Resolving the limitations of a low-dimensional action space, it models the joint distribution of a high-dimensional treatment $A$ as a mixture over a latent substitute confounder similar to \eqref{eq:learn-from-a}. By assuming individual treatments are conditionally independent given this latent factor, learning this mixture
\textcolor{black}{can help recover a substitute confounder,
potentially enabling downstream causal estimation
under additional identifiability assumptions}.

\emph{Proximal causal learning} is another related and promising framework \cite{proximal} that identifies causal effects using multiple {\em proxies}
rather than multiple causes. This structural setup is highly reminiscent of multi-view learning models, where independent views conditioned on a latent state enable efficient moment-based recovery \cite{anandkumar2012method}. This connection naturally motivates extending efficient tensor and moment-based methods from multi-view learning to proximal causal learning--this is precisely what is done \textcolor{black}{in recent work} \cite{saha}. 


{
\section*{Opportunities}
Despite the progress discussed, a lot remains to be done. For example, there are restrictions on the methods discussed for mixture learning \textcolor{black}{in terms of finite latent types, strong separability, high dimensionality, or specific types of conditional independence}. To begin with, \textcolor{black}{making progress on empirical, data-driven validation of the conditions required for the validity of each method} is an important open direction. Beyond that, developing less restrictive mixture learning \textcolor{black}{methods applicable to causal estimation} is definitely an important direction to be explored. For example, \textcolor{black}{extensions of mixture learning for exponential families to continuous or countably infinite latent spaces, replacing hard clustering with estimation of the latent mixing distribution, are important directions.} \textcolor{black}{Extending causal estimation in the presence of overlapping mixture components, potentially even in the absence of mixture identifiability, will also be an exciting direction to work towards.} It is worth commenting that the mixture-learning perspective complements rather than replaces approaches such as instrumental variables. Its utility depends on whether the latent heterogeneity admits a learnable mixture representation, while alternative approaches may be more natural when suitable instruments are present. This synergy between causal inference and mixture learning points toward scalable methods for high-dimensional settings, while opening new questions at the interface of the two disciplines.
}


\newpage
\bibliographystyle{IEEEtran}
\bibliography{ref}
\end{document}